\documentclass[preprint,12pt,authoryear]{elsarticle}

\usepackage{amssymb}
\usepackage{amsmath}
\usepackage{booktabs}
\usepackage{graphicx}
\usepackage{xspace}
\usepackage{lineno}
\usepackage{multirow}
\usepackage{algorithm}
\usepackage{algorithmic}
\usepackage{hyperref}
\usepackage{xcolor}
\def\method{Z(PM)$^2$\xspace}
\def\eg{e.g.,\xspace}

\journal{ISPRS Journal of Photogrammetry and Remote Sensing}

\begin{document}

\begin{frontmatter}



\title{\method: Zero-Shot Polygon Matching with Pre-trained Models for Pose Estimation and Polygon Cloud from Challenging Stereo Images} 
\author[label1]{Chang Li\corref{cor1}} 
\author[label1]{Xingtao Peng}
\affiliation[label1]{organization={Key Laboratory for Geographical Process Analysis \& Simulation of Hubei Province, and College of Urban and Environmental Sciences},
addressline={Central China Normal University}, city={Wuhan}, postcode={430000}, state={HuBei}, country={China}}
\cortext[cor1]{Corresponding author. E-mail: lcshaka@126.com \&\& lichang@ccnu.edu.cn}
\begin{abstract}
While stereo matching has achieved maturity for 0D point and 1D line primitives, establishing correspondences for 2D polygons remains largely unexplored due to challenges including disparity discontinuity, scale variation, training dependency, and poor generalization, limiting its potential for downstream tasks such as pose estimation and 3D reconstruction. To address these issues, we are the first to propose a Zero-shot Polygon Matching paradigm with Pre-trained Models (i.e., \method), which combines learned features and handcrafted geometric constraints through plug-and-play modules, extending matching from 0D/1D primitives to 2D polygons. The pipeline comprises three core stages: Firstly, detector leverages the pre-trained segment anything model to vectorize segmentation masks into graph-structured polygons integrating both geometry and texture; Secondly, global matcher uses bidirectional-pyramid and multi-geometric constraints to handle global viewpoint variation; Thirdly, local matcher leverages local-holistic bipartite graph optimization to resolve local disparity discontinuity and topological inconsistency. Moreover, we further develop polygon-matching-guided pose estimation using correspondences to obtain well-distributed, low-redundancy homologous points, and pioneer the polygon cloud concept with optimal surface generation method, producing structurally complete and semantically rich 3D representations beyond point and line clouds. Since no existing polygon matching methods from stereo imagery have been available for direct comparison, we selected state-of-the-art (SoTA) methods close to this task as alternative baselines. Extensive experiments on five challenging datasets (ISPRS, KITTI, ScanNet, SceneFlow, DTU) show \method achieves a 68.60\% matching area score, outperforming MESA by approximately 32\% and ranking first in area-level pose estimation, with competitive speed and strong zero-shot generalization without any training requirement.
\end{abstract}

\begin{keyword}
Polygon matching \sep Stereo matching \sep Pre-trained model \sep Disparity discontinuity \sep Zero-shot
\end{keyword}

\end{frontmatter}



\section{Introduction}
\label{sec:intro_new}
\begin{figure}[h!]
    \centering
    \includegraphics[width=\linewidth]
    {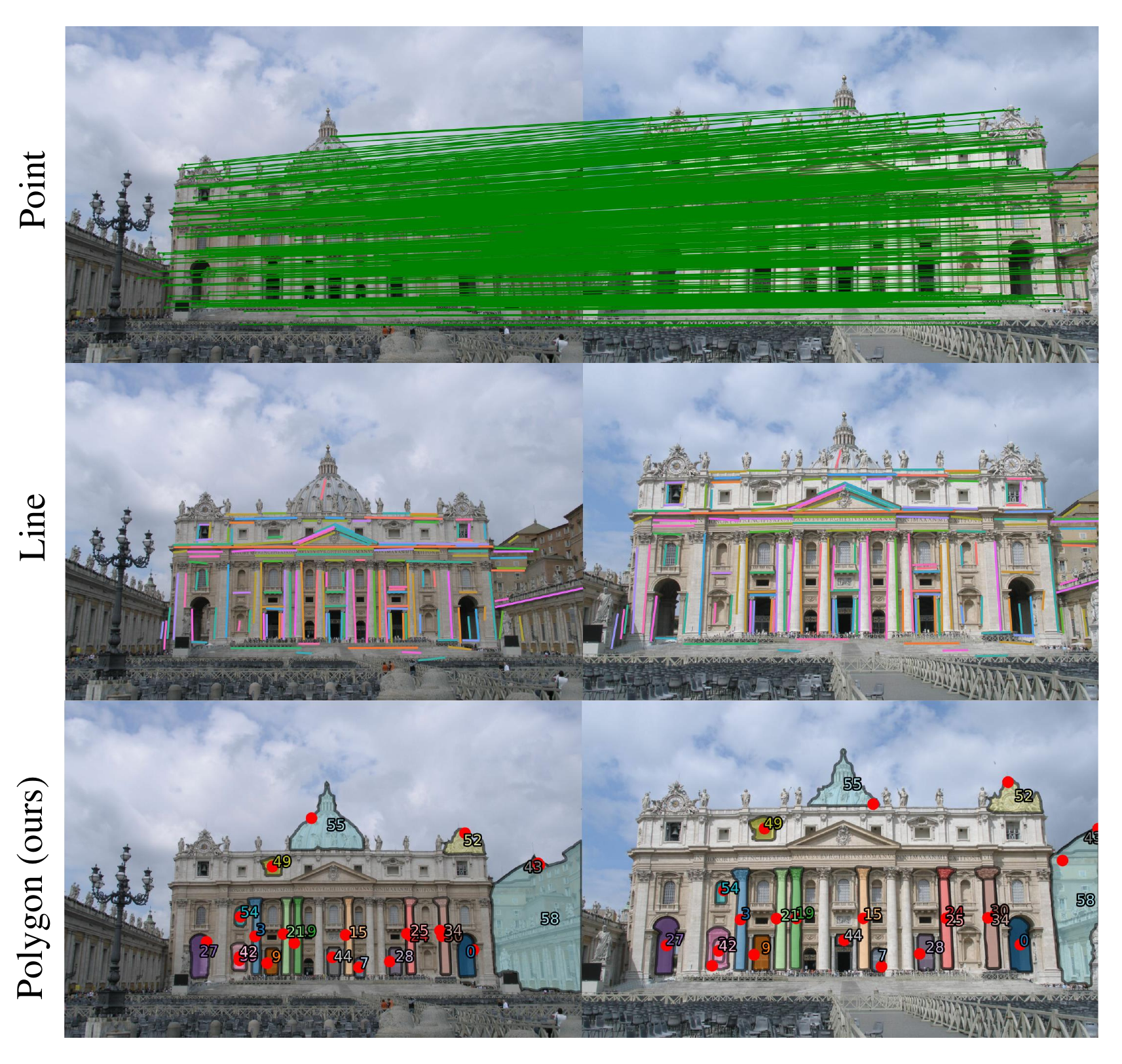}
    \caption{A visual representation of the different types of matching primitives (points, lines, polygons).}
    \label{fig:1}
\end{figure}

In photogrammetry, remote sensing and computer vision, stereo image matching establishes correspondences between images captured from different viewpoints \citep{Zhang_Zhu_Li_Lu_Ma_2025}, and achieving high accuracy and robust correspondences in this process is vital for a wide range of downstream tasks \citep{xuLocalFeatureMatching2024a}, such as multimodal image matching \citep{jiangReviewMultimodalImage2021}, data fusion, and pose estimation.
Currently, handcrafted feature-based and deep-learning-based methods have achieved remarkable progress in point and line correspondence: (1) Point matching aims to find corresponding individual points between images.
Classic algorithms such as SIFT \citep{loweObjectRecognitionLocal1999} have been augmented and often surpassed by deep learning-based methods such as SuperPoint \citep{detoneSuperPointSelfsupervisedInterest2018} and local feature transformer (LoFTR) \citep{sunLoFTRDetectorfreeLocal2021} that learn features directly from data, achieving superior matching accuracy and robustness.
(2) Line matching \citep{linComprehensiveReviewImage2024} focuses on establishing correspondences between line segments and provides structural information; (3) Hybrid point-line matching, GlueStick \citep{pautratGlueStickRobustImage2023} unifies point and line into a single wireframe structure to achieve more accurate and reliable point-line matching results.
While point and line matching have achieved significant progress, higher-level structural feature matching such as polygon matching remains unexplored from stereo images.
Therefore, as shown in Fig.~\ref{fig:1}, this paper extends the concept of image matching (i.e., point and line) to a higher dimension: polygon matching (i.e., face), which possesses a clear geometric structure and semantic meaning \citep{dengMultimodalPlaneInstance2025}.

\begin{figure}[h!]
    \centering
    \includegraphics[width=\linewidth]{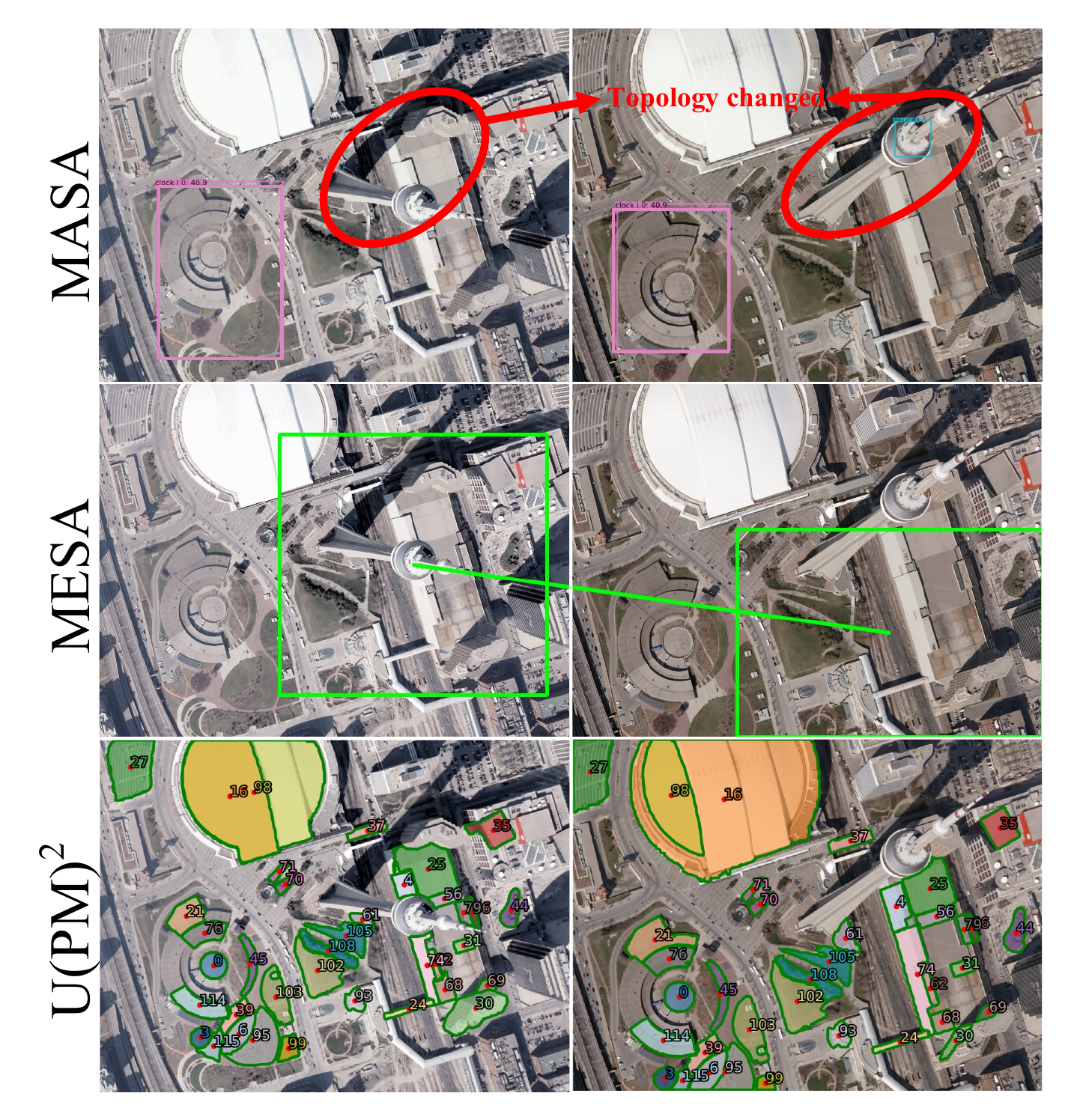}
    \caption{\textbf{\method vs. MASA and MESA in visual comparison.} The locations indicated by the rounded boxes are where the topological relationships have changed. MASA fails to detect polygons or areas because it was designed for different tasks. MESA achieves lower area overlap ratios and higher mismatch rates. In contrast, \method attains the highest number of matches with the finest-grained accuracy (down to individual building-level accuracy). It also shows a difficult case (marked with a red-circle) where topological relationship changes (from adjacency to separation) due to large viewpoint shifts, causing severe geometric distortion, occlusion and spatial conflicts.}
    \label{rw:1}
    \end{figure}

Polygons offer a more comprehensive representation of scene structure than individual points and lines, so they are particularly crucial in scenarios demanding high-level scene understanding \citep{jiaoLDPolyLatentDiffusion2025}, such as urban reconstruction from aerial imagery or detailed 3D modeling for augmented reality.
Polygon matching has not been explored yet, and faces some key challenges: (1) large-format imagery matching demands robust solutions to large-scale variation for remote sensing imagery, which includes adapting to both the sheer size of the images and the significant scale differences between corresponding polygons across stereo pairs \citep{liFusionAerialMMS2023}; (2) local disparity discontinuities (Fig.~\ref{rw:1}) lead to local topological inconsistencies, so dedicated algorithms are required to overcome this contradiction; (3) generalization should be ensured in zero-shot (i.e., training-free) polygon matching.

To address the aforementioned challenges in stereo polygon matching, we propose \method, the first zero-shot polygon matching framework with pre-trained models.
\method aims to establish fine-grained polygon-to-polygon correspondences in stereo images through a multi-stage pipeline designed to tackle scale variation, search efficiency, local disparity discontinuities and generalization with training-free strategy. The pipeline includes the following: Firstly, the feature points and vector polygons are detected in images by the detector with pre-trained models, laying the foundation for subsequent matching.
Secondly, to address the challenge of scale variation inherent in large-format images, the global matcher leverages the bidirectional-pyramid matching (BPM) strategy to generate the adaptive search window, which progressively narrows the search space in the image pyramid, significantly reducing computational cost while maintaining matching accuracy. The reliable correspondences established at this easy stage serve as geometric anchors that constrain and quality-control the subsequent local matching.
Thirdly, we refine the matching results by local-joint geometry and multi-feature matching strategy (LoJoGM) with the local-holistic bipartite graph optimization.
This step is crucial for mitigating local disparity discontinuities, as it considers the local relationships between polygons and polygons from stereo images, and addresses topological inconsistencies in the target as the viewpoint changes. As the final quality-control stage, LoJoGM re-matches the hard cases that the global stage could not resolve and rejects only those correspondences for which no candidate yields a globally consistent assignment, thereby controlling error propagation across the whole pipeline.
By combining automatically learned features with handcrafted features and an optimized matching strategy, \method offers a zero-shot and generalized solution for stereo polygon matching without any training requirement.

In summary, this paper makes the following key contributions to the field of stereo image matching:
\begin{enumerate}[(1)]
\item We are the first to propose a zero-shot polygon matching paradigm \method that combines automatically learned features from pre-trained models with handcrafted geometric constraints.
Moreover, in our paradigm, any module (e.g., pre-trained model) can be replaced by more advanced algorithms.
This paradigm follows a progressive quality-control design, in which each stage refines and corrects the residual errors left by the previous one, so that mistakes are mitigated stage by stage rather than propagated unconditionally.
This work advances the field of image matching, upgrading from 0/1D (point/line) primitives to 2D (polygon), thus providing a zero-shot, training-free solution with strong generalization for correspondence problems.
\item We propose a global matcher based on BPM and multi-geometric constraints with pre-trained LoFTR, which addresses the critical challenge of viewpoint and scale variation in large-format imagery (e.g., remote sensing image), enabling efficient and accurate polygon matching in challenging scenarios.
\item We propose a local-joint geometry and multi-feature matching strategy, i.e., LoJoGM, which effectively mitigates local disparity discontinuities by considering both graph and texture features.
Theoretically, we propose a local-holistic bipartite graph optimization (LHBGO) paradigm that bypasses explicit topological and disparity relationships to alleviate ill-posed matching and produce robust, mutually non-overlapping polygon correspondences, providing well-distributed homologous points for improving the accuracy of pose estimation and densifying matched polygons for polygon cloud construction.
Specifically, local matching re-matches the polygons that failed in global matching to recover more reliable correspondences, and partially overcomes the topological conflicts and abrupt disparity changes that violate the global coplanarity assumption.
\item We establish a dedicated evaluation metric specifically designed for polygon matching tasks to address the inadequacy of existing metrics and demonstrate the state-of-the-art (SoTA) performance of \method against existing methods using this metric as well as other commonly employed matching evaluation metrics.
Unlike traditional point/line metrics or simple IoU that often overlook over-coverage and topological inconsistencies, our proposed metrics symmetrically penalize shape deviations and redundant matches to ensure a more rigorous assessment.

\item We propose polygon cloud as an innovative 3D representation primitive reconstructed from stereo polygon matching, together with an optimal polygon surface generation algorithm combining polygon diffusion approximation and polygon edges of projection surface. Unlike discrete point clouds and open line clouds, polygon cloud lifts the primitive to 2D closed surface-conforming patches, offering higher dimensionality, semantic richness, structural completeness, and surface fidelity. To our knowledge, defining and generating polygon clouds directly from stereo polygon matching has not been reported before.

\end{enumerate}
\section{Related work}
\label{sec/2_related_work}
\subsection{Stereo matching}
Stereo matching can be broadly classified into area-based matching (ABM) and feature-based matching (FBM) \citep{gengFeaturebasedMultimodalRemote2025}.
ABM approaches minimize a matching cost function to reduce mismatches or aggregate matching costs within local pixels as the matching criterion \citep{luoFastNormalizedCrosscorrelation2010}.
Traditional FBM approaches match images by extracting and computing feature descriptors.
Deep learning-based FBM \citep{laiMatchDetCollaborativeFramework2024,Zhang_2025_CVPR} further advances this field by training neural networks to automatically learn feature representations.
It enables precise matching in complex scenes, low-texture regions, and under occlusions. Both ABM and FBM are suitable for feature-rich scenes with large illumination variations, but FBM is more computationally efficient than ABM.
Notably, learning-based matchers such as SuperGlue \citep{sarlinSuperGlueLearningFeature2020} and LightGlue \citep{lindenbergerLightGlueLocalFeature2023} cast matching as an optimal transport (OT) problem, performing a global soft assignment over point-level descriptors. In contrast, the LHBGO mechanism in \method operates on polygon-level primitives and yields a hard one-to-one assignment confined to local subgraphs anchored by reliable matches, requiring no training.

Distinct from these conventional approaches, \method pioneers a paradigm shift from 0D/1D primitives to 2D polygon matching, offering a robust zero-shot solution for high-level structural correspondence by uniting the robustness of learned features with the precision of handcrafted geometric constraints.
By replacing computationally intensive dense pixel matching with sparse, polygon-level structural alignment, \method significantly reduces computational overhead, offering an efficient solution for stereo matching.\method performs inference directly on new scenes without any task-specific training or fine-tuning, whereas many learning-based matchers require days or even weeks of supervised training on large-scale labeled data before deployment.

\subsection{Polygon matching}
\label{sec:related_shape}
Several lines of research are related to but distinct from polygon matching in stereo images. We review them in order of increasing relevance and clarify the boundary of our task definition.

\textbf{Shape matching and retrieval.}
Classical shape matching methods, such as Shape Context \citep{belongieShapeMatchingObject2002} and Inner Distance Shape Context (IDSC) \citep{lingShapeClassificationUsing2007}, measure shape similarity by computing contour descriptors from known vector templates.
These methods operate in a one-to-many retrieval setting with pre-stored vector boundaries, relying solely on geometric features without incorporating texture information. Their outputs (similarity scores or template rankings) are designed for shape recognition and retrieval, and do not interface with downstream geometric tasks such as pose estimation or 3D reconstruction.

\textbf{Region detection and description.}
Region detection methods such as MSER \citep{matasRobustWidebaselineStereo2004} extract stable extremal regions approximated as elliptical windows, producing point-level (0D) correspondences rather than polygon-to-polygon matches.
Modern deep feature methods such as D2-Net \citep{dusmanuD2NetTrainableCNN2019} and R2D2 \citep{revaudR2D2ReliableRepeatable2019} detect and describe keypoints at patch centers, and their output remains point-level matching despite being region-oriented.

\textbf{Area-assisted point matching and object tracking.}
Although polygon matching from stereo imagery has not been reported, a few related works exist that explore similar problems.
Matching everything by segmenting anything (MESA) \citep{zhangMESAMatchingEverything2024} and semantic and geometry area matching (SGAM) \citep{zhangSearchingAreaPoint2024a} utilize segmented regions merely as auxiliary constraints to refine pixel-level or point-level correspondences.
By definition, they treat areas as homogeneous filters rather than independent matching primitives, thereby ignoring the explicit vector topology and structural alignment required for polygon matching.
Matching anything by segmenting anything (MASA) \citep{liMatchingAnythingSegmenting2024} focuses on object association across continuous video frames.
They rely heavily on temporal continuity, which makes them fundamentally ill-suited for stereo matching tasks.
Unlike stereo pairs, which exhibit significant disparity discontinuities and scale variations due to large baselines, tracking methods are not designed to handle the geometric deformations inherent in non-continuous stereo views.
As shown in Fig.~\ref{rw:1}, when applied to stereo polygon matching, both methods face notable challenges.
Specifically, MASA is suitable for target tracking tasks but not for polygon matching, particularly in complex real-world scenes.
MESA exhibits lower area overlap ratios and higher mismatch rates, often failing to achieve building-level accuracy, and shows reduced performance under significant viewpoint changes.
These characteristics arise from the design scope of each method:
(1) MASA relies on the generalization of its object detection and tracking modules, which may produce mismatches on untrained targets (\eg aerial imagery);
(2) MESA and SGAM are designed for point matching rather than polygons and depend on robust spatial (i.e., graphic) relationships, yet disparity abrupt changes induce spatial inconsistencies, leading to matching failures;
and (3) the inherent sensitivity of both methods to large viewpoint variations.

In contrast to the above research, the polygon matching proposed in this paper is defined as establishing one-to-one correspondences of closed 2D polygons from rasterized stereo imagery, which is fundamentally different from the aforementioned approaches in three respects: (1) we operate on rasterized stereo images with significant geometric deformations, rather than pre-stored vector templates; (2) our matching output is a set of geometrically precise polygon-to-polygon correspondences with explicit vector topology, rather than point-level matches or similarity rankings; and (3) the matched polygons can directly interface with downstream tasks in computer vision and remote sensing, including pose estimation and 3D reconstruction.

\subsection{Downstream tasks of matching}
\label{sec:downstream_rw}

\subsubsection{Pose estimation from correspondences}
Camera pose estimation from image correspondences is a well-established task in photogrammetry and computer vision.
Given a set of point correspondences between two views, the essential matrix $\mathbf{E}$ encodes the relative rotation and translation.
The five-point algorithm combined with RANSAC provides a robust estimation pipeline.
Recent learning-based matchers such as SuperGlue \citep{sarlinSuperGlueLearningFeature2020} and LoFTR \citep{sunLoFTRDetectorfreeLocal2021} improve the quality of input correspondences, thereby enhancing pose accuracy.
Area-level pose estimation methods leverage region or area correspondences (rather than isolated point features) to constrain pose recovery. MESA \citep{zhangMESAMatchingEverything2024} represents the current SoTA in this category, having demonstrated superior performance. Inspired by the above advancements, we extend pose estimation from point-level to polygon-level correspondences, validating that polygon matching provides geometrically accurate inputs for pose recovery.

\subsubsection{From point and line cloud to polygon cloud}
Point clouds acquired by LiDAR or reconstructed via multi-view stereo (MVS) \citep{schonbergerStructurefromMotionRevisited2016} represent scenes as discrete 0D samples that lack explicit boundary and semantic structure.
Line clouds \citep{pautratGlueStickRobustImage2023, weiELSREfficientLine2022} extend representation to 1D segments extracted and triangulated from images, yet line segments are open primitives prone to fragmentation and do not convey area or instance information.
Both representations fall short of capturing the enclosed, semantically meaningful surfaces of individual objects.

To our knowledge, polygon cloud the first work to define and generate polygon clouds from stereo polygon matching results. Polygon cloud is a set of closed polygonal meshes that conform to the actual 3D object surfaces.
Each element is a surface-conforming polygonal mesh that follows the true 3D object curvature, composed of triangular facets from the reconstructed model, capable of representing complex surface geometry.

Compared with point clouds and line clouds, polygon cloud offers three fundamental advances in representational completeness: (1) Structural completeness. Each polygon forms a topologically closed boundary that unambiguously delineates an object region, as opposed to the unstructured and boundary-free nature of point clouds; (2) Semantic richness. Each closed polygon inherits instance-level identity from the matching stage without requiring additional 3D semantic annotation; and (3) Surface fidelity. Polygon boundaries trace geodesic-like paths along the reconstructed mesh surface, preserving 3D curvature, area, and surface normal information that 0D and 1D representations inherently do not encode.

\section{Methodology}
\label{sec:methodology}
\begin{figure}[h!t]
    \centering
    \includegraphics[width=\linewidth]
    {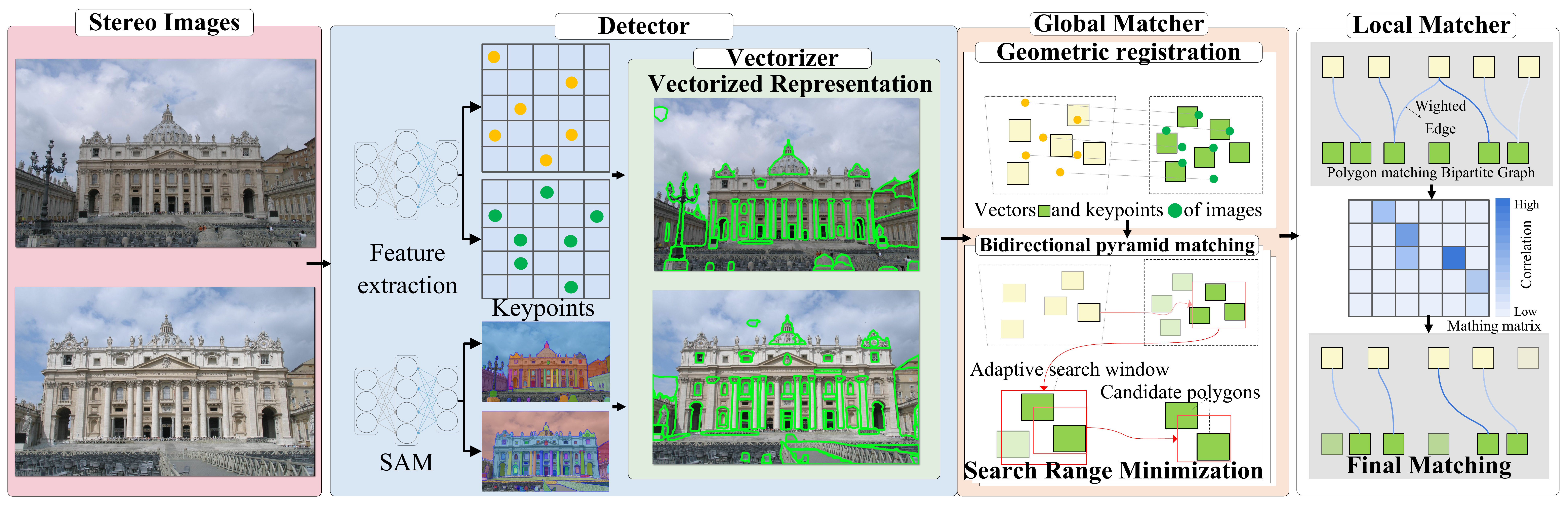}
    \caption{\textbf{Overview of \method}.
	\textbf{(1)} Detector jointly performs polygon and feature point detection from stereo images to construct polygons with graphic structure. \textbf{(2)} Global matcher matches feature points extracted by the Detector, establishing reliable correspondences and global geometric constraints. \textbf{(3)} Local matcher eliminates ambiguous matches through the geometry and multi-feature matching strategy to solve bipartite graph matching, ultimately optimizing polygon matching.}

\label{fig_main}
\end{figure}

Fig.~\ref{fig_main} illustrates the architecture of \method.
(1) The detector constructs the foundational primitives by abstracting pixel-level masks from the pre-trained SAM into graph-structured polygons, establishing a dual representation of skeletal topology and internal texture; quality is controlled here by requiring only cross-view repeatable polygons rather than absolutely correct segmentation;
(2) The global matcher efficiently anchors global geometric constraints using a bidirectional-pyramid strategy, which progressively narrows the search space from a coarse global view to fine-grained local regions; quality is controlled by establishing globally reliable correspondences as geometric anchors that constrain the next stage;
(3) The local matcher resolves the ill-posed problem of topological inconsistency through the local-joint geometry and multi-feature (LoJoGM) strategy, enforcing optimal many-to-many alignment via local-holistic bipartite graph optimization. Quality is controlled by re-matching the polygons that the global stage failed to match and rejecting candidates without a globally consistent assignment.
Rather than treating the three stages as independent steps, \method organizes them into an easy-to-hard, progressive quality-control pipeline, in which each stage establishes reliable evidence for the next and corrects the residual errors left by the previous one, so that errors are suppressed stage by stage instead of being propagated unconditionally.

\subsection{Definition of polygon matching }
\begin{figure}
    \centering
    \includegraphics[width=\linewidth]{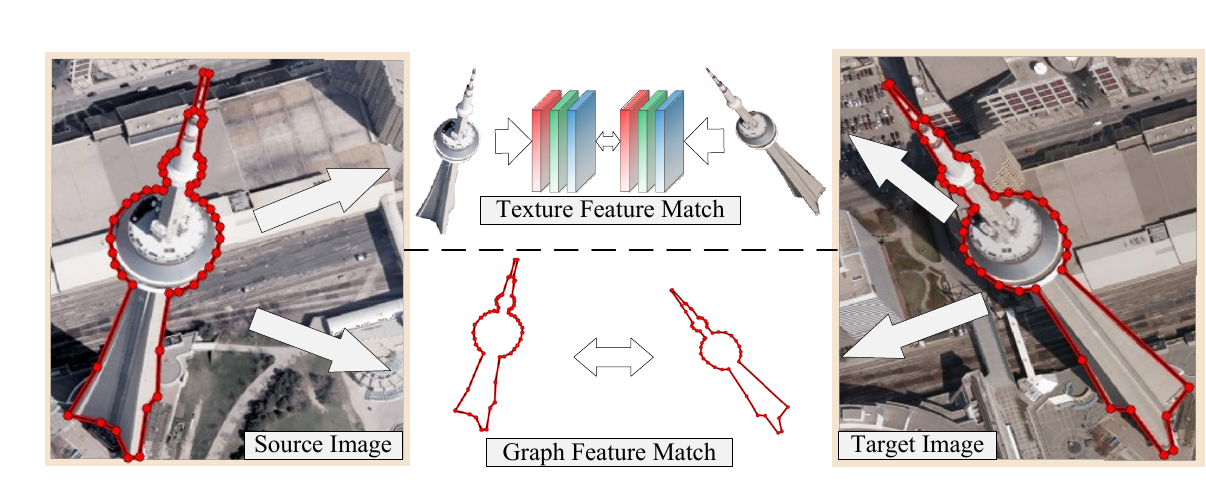}
    \caption{Illustration of polygon matching. The example images shown in the figure are also considered a typical scenario of topological inconsistency, which is difficult to match.}
    \label{fig:problem_def}
\end{figure}

As shown in Fig.~\ref{fig:problem_def}, the objective of polygon matching is not only to extract the polygons but also to match both their graph and the texture features.
Let $\mathcal{A}=\{P_i\}_{i=1}^m$ denote the set of source polygons and $\mathcal{B}=\{P_j^\prime\}_{j=1}^n$ denote the set of target polygons.
Each polygon $P_i$ (or $P_j^\prime$) is represented by its set of geometric attributes $G_i$ (or $G_j^\prime$) and features $F_i$ (or $F_j^\prime$):
\begin{equation}
    {{{P}}_i} = \{ {G_i},{F_i}\}
\end{equation}
We aim to find an optimal matching between polygons in $\mathcal{A}$ and $\mathcal{B}$, subject to geometric constraints and features as shown in Fig.~\ref{fig:problem_def}.
The matching result is represented by a set of binary relation matrices:
\begin{equation}
    {M_{ij}} = \mathbb{I}({G_i} \cong {G'_j}) \cdot \mathbb{I}({F_i} \cong {F'_j}),\quad \forall {P_i} = \{ {G_i},{F_i}\} \in \mathcal{A},{P'_j} = \{ {G'_j},{F'_j}\} \in \mathcal{B}
\end{equation}

where $\mathbf {M}_{ij}$ represents the element in the matching matrix $\mathbf {M}$ (indicating whether $P_i$ matches $P_j^\prime$).
$\cong$ denotes similarity satisfies a threshold.
$\mathbb{I}$ is indicator function, the value is 1 if the condition inside the parentheses (\eg $G_i \cong G^\prime_j$ ) is true; otherwise, it is 0.

\subsection{Detector}
The detector includes polygon detection and feature point detection.
Polygon detection consists of the segmentor and the vectorizer: a pre-trained SAM is first used to perform zero-shot instance segmentation, producing object-level masks and bounding boxes; then, a vectorization step converts each mask contour into a polygonal representation.
Next, the Douglas–Peucker algorithm is applied to simplify the vector nodes and remove edge noise due to raster-vector conversion.
Finally, the graph $G$ is constructed based on the nodes of the polygon:
\begin{equation}
    G=\left(V,E\right)
\end{equation}
where $V$ represents the set of all vertices (i.e., the set of polygon vertices), and $E$ denotes the set of undirected edges $e_i$, corresponding to the polygon edges.
For point detection, \method integrates both handcrafted features (e.g., SIFT) and learned features (e.g., SuperPoint, LoFTR, LightGlue) to improve robustness and enable zero-shot polygon matching.
As the starting point of stage-wise quality control, the detector only needs to provide repeatable SAM features across stereo views, which is the very prerequisite of stereo feature matching (this stage is also termed sparse matching in photogrammetry and computer vision, serving SfM and aerial triangulation); hence dense polygon features are unnecessary. Moreover, whether the segmentation is over- or under-partitioned does not affect matching as long as the polygons extracted from the left and right images are mutually repeatable (e.g., both over- or both under-segmented), since the partitioning granularity is itself partly subjective and only cross-view consistency is required.

\subsection{Global matcher}
The global matcher performs coarse matching: it first establishes globally reliable correspondences under viewpoint and scale variation, which serve as geometric anchors that constrain and quality-control the subsequent local matching.
The global matcher contains the following key modules:
(1) The initial matching module establishes fundamental correspondences between stereo image pairs by estimating a global homography matrix through a pre-trained LoFTR or others.
(2) The bidirectional pyramid matching module dynamically optimizes search ranges across multi-scale pyramid features to balance computational efficiency and matching completeness.
(3) Geometric rectification compensates distortions between polygon regions.\\
\subsubsection{Initial matching}
First, we use a feature matcher to get initial matching on the feature points extracted by the detector.
This results in the fundamental matrix $\mathbf{F}$ and the homography matrix $\mathbf{H}$, which filters mismatches by marginalizing sample consensus++ (MAGSAC++) \citep{barathMAGSACFastReliable2020a}, providing an approximate spatial correspondence for source polygons.
Due to the significant viewpoint changes in stereo images, homography estimation is not always accurate.
Therefore, before calculating the homography matches for the computed matched points, we first estimate the fundamental matrix, compute the epipolar lines, and then retain only the matched feature points sufficiently close to epipolar lines while removing outliers far from these lines.
It is important to note that the discarded feature points are still retained but excluded from the homography calculation.
This method allows us to preserve many feature points for further matching in the subsequent steps.\\
\begin{figure}[!h]
    \centering
    \includegraphics[width=0.8\linewidth]{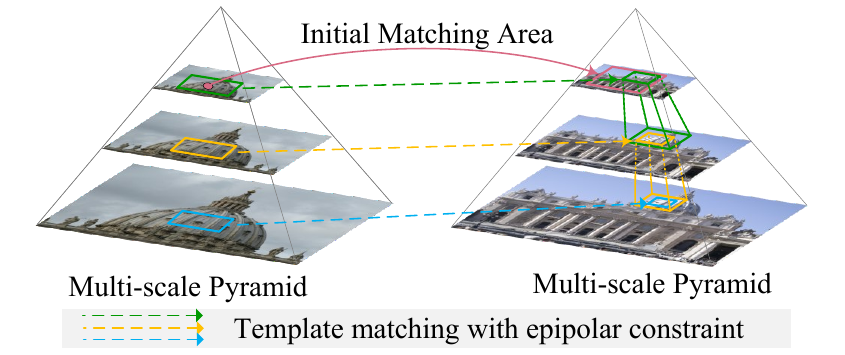}
    \caption{\textbf{Bidirectional-pyramid matching. }The bidirectional pyramid progressively narrows the search region from the top-level initial area toward lower levels to establish geometric constraints for subsequent local matching.}
    \label{method_pyramid}
\end{figure}

\subsubsection{Bidirectional-pyramid matching}
\label{sec:BPM}
To address the adaptation of the search range for matching images of varying sizes and to improve matching efficiency and accuracy, bidirectional-pyramid matching (BPM) is proposed.
Fig.~\ref{method_pyramid} shows that this strategy reduces computational complexity by combining multi-stage downsampling and adaptive search window and improves the matching accuracy by utilizing epipolar line constraints and template matching detailed in Algorithm \ref{alg:1}.

\begin{algorithm}
    \caption{Bidirectional-pyramid matching}
    \label{alg:1}
    \begin{algorithmic}[1]
    \STATE Construct search window $W_\phi$ with $c_i$ and $\phi$
    \STATE Crop template $A$ from layer $k$ of image $X$ using window $W_\phi$
    \IF{$k == 0$}
        \STATE Set $\tau \gets 25$
        \STATE Construct search window $W_\tau$ with center $c'_i$ and size $\tau$
    \ELSE
        \STATE Map final search window $W_\phi$ from layer $k-1$ to current layer as $W_\tau$
    \ENDIF
    \STATE Crop search area $S$ from layer $K$ of image $Y$ using window $W_\tau$
    \STATE Perform template matching between $A$ and $S$ to find most correlated point $c''$
    \STATE Construct new search window $W_\phi$ centered at $c''$ with size $\phi$
    \end{algorithmic}
    \end{algorithm}

First, image Gaussian pyramids are created for the stereo images with a downsampling factor of 1/3.
Downsampling stops when the length or width of the image at the bottom of the Gaussian image pyramid does not exceed 200 pixels.
For an image size of $W\times{H}$, $n$ layers of pyramids, $\text{min}\{\ W/3^n,\ H/3^n\}<200$, so the time complexity of subsequent matching on large-format images is reduced.
At the top level of the pyramid, the center $c_i$ of the source polygon is transformed to $c_i^\prime$ in the target image using the homography matrix.
A square window $W_\tau$ of size $\tau$ is then centered at $c_i^\prime$, and a smaller window $W_\phi$ of size $\phi$ is centered at $c_i$.

Next, $W_\phi$ is used as a template to perform template matching with $W_\tau$, identifying the most similar region $W_\phi^\prime$ centered at $c_i^{\prime\prime}$ in the target image:
\begin{equation}
{W^\prime_{\phi }}=\left\{ (x,y)\left| |x-x_{c^{\prime\prime}_i}|\le \frac{\phi }{2} ,|y-{{y}_{c^{\prime\prime}_i}}|\le \frac{\phi }{2} \right. \right\}
\end{equation}
where $x$, $y$ denote the coordinates of points.
Subsequently, the candidate matching region $W_\phi^\prime$ is then filtered using epipolar constraints.
The epipolar constraint is used to evaluate whether the center points $c^{\prime\prime}$ of the matching region are within a range $\epsilon$ above or below the epipolar line based on the fundamental matrix $\mathbf{F}$:
\begin{equation}
    \left| \boldsymbol{p}_{\text{right}}^{\top}\mathbf{F}{\boldsymbol{p}_{\text{left}}} \right|\le \epsilon
\end{equation}
where the point $c$ and $c^{\prime\prime}$ are first homogenized into $\boldsymbol {p}_\text{left}=\left(x_c,y_c,1\right)^\top$ and $\boldsymbol {p}_\text{right}=\left(x_{c^{\prime\prime}},y_{c^{\prime\prime}},1\right)^\top$.

Finally, $W_\phi^\prime$ is projected onto the lower levels of the pyramid, and the process is repeated.
Typically, passing through four pyramid levels is sufficient to narrow down the search to an accurate range $W_{\mathrm{final}}$.
The target polygons within the $W_{\mathrm{final}}$ are considered the candidate matching polygons for the corresponding source polygon.
This process is repeated for all source polygons $\mathcal{A}=\{{P}_i\}_{i=1}^m$ and target polygons $\mathcal{B}=\{{P}_j^\prime\}_{j=1}^n$ to obtain the coarse matching result $\mathbf{M}\in\{0,1\}^{m\times n}$.
$m$ and $n$ represent the number of source polygons and target polygons, respectively.
$\mathbf{M}_{ij}=1$ when ${P}_i$ matches ${P}_j$.
$\mathcal{A}$ and $\mathcal{B}$ are many-to-many relationships, and further refinement of the matching is required in the local matcher to optimize the matches to one-to-one relationships and to eliminate some false matches.

\subsection{Local matcher}
\begin{figure}[!ht]
    \centering
    \includegraphics[width=\linewidth]{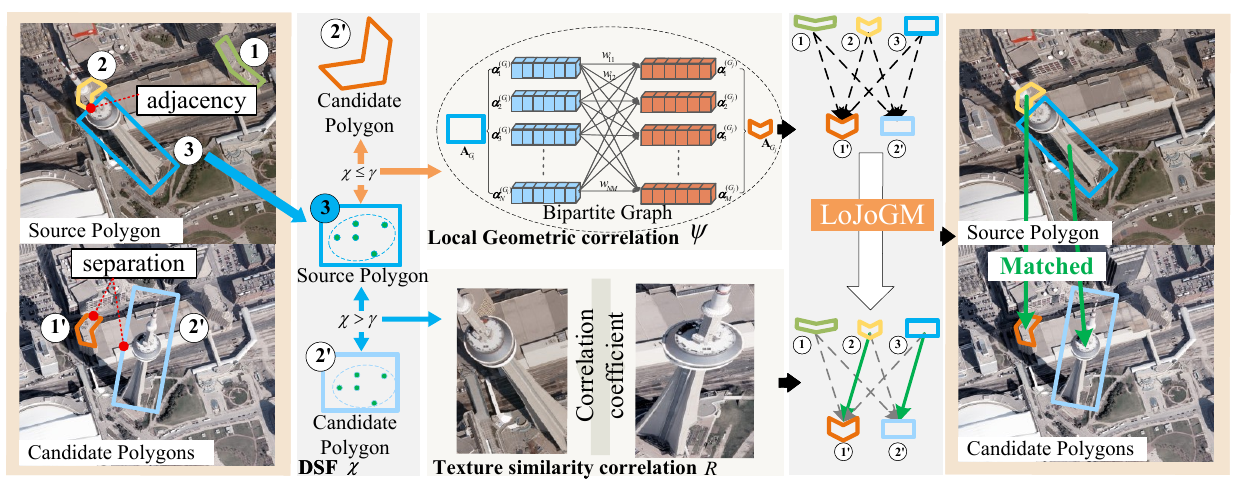}
    \caption{ \textbf{The proposed local matcher.} It addresses matching failures caused by local deformations and topological inconsistencies.
    The essential reason is the viewpoint variation causes disparity variation, then disparity variation in turn leads to topological inconsistency (e.g., the relative position relationship between the box and the polygonal area).
    To solve the above-mentioned issue, the local matching is proposed by joint local geometric and texture correlations.}
    \label{method_local}
\end{figure}

To address the legacy issues of global matching, local matcher is used to refine correspondences within the local-identified candidate polygons (Fig.~\ref{method_local}).
As the final fine-matching stage, the local matcher re-matches the polygons that the coarse global stage failed to match, exploiting their local relationships for a holistic optimal assignment, thereby controlling error propagation across the whole pipeline.
The process begins by evaluating the Deep-Spectral factor (DSF), defined as the count of homologous points matched by a pre-trained matcher (e.g., LoFTR) between source and candidate polygons.
This metric dictates a dual-branch correlation strategy: (1) If the DSF satisfies the requirements for homography estimation, the candidate polygon undergoes local geometric rectification, followed by the computation of local geometric correlation (i.e., graph similarity); (2) otherwise, texture similarity correlation is computed directly.
In either case, the resulting similarity score is converted into a standardized matching cost.
Finally, the LoJoGM module employs local-holistic bipartite graph optimization to minimize the total cost to determine optimal polygon correspondences, ensuring robust matching even with limited geometric constraints.\\

\subsubsection{Local geometric correlation}
In local matching, local geometric distortions pose a significant obstacle to accurate polygon matching.
Even though preliminary registration may have been performed in the global matching, geometric distortions still exist locally due to the registration errors being globally optimal.
To address this issue, we propose local geometric correlation, which is designed to further reduce local geometric distortions or ambiguities in the polygons to be matched.
First, if the DSF satisfies the conditions for a local homography transformation, the polygon is corrected locally.
Then, the local geometric correlation is described by both their area discrepancy and the geometric discrepancy.

For the local geometric discrepancy $\beta$, we compute it using the graph-matching algorithm \citep{kioucheMaximumDiversitybasedPath2021}.
It represents each node $v_i$ in all $N$ nodes of the graph (i.e., polygons) using its top-$k$ neighboring nodes embedding $\boldsymbol{\alpha}_i$.
This enables the conversion of the graph into a vector embedding $\boldsymbol{A}=\left(\boldsymbol{\alpha}_\mathbf{1},...,\boldsymbol{\alpha}_{\boldsymbol{N}}\right)^\top$
\begin{equation}
    {{\boldsymbol\alpha }_{i}}=\left( e_{1}^{i},e_{2}^{i},\ldots ,e_{k}^{i} \right),\quad \text{where } 1\le k\le N-1
\end{equation}
\begin{equation}
    e_{k}^{i}=\frac{d({{v}_{i}},{{v}_{\mathbb{E}\left( {{v}_{i}},k \right)}})}{\sum\limits_{l=i}^{\mathbb{E}\left( {{v}_{i}},k \right)-1}{d}({{v}_{l}},{{v}_{l+1}})}\sum\limits_{l=i}^{\mathbb{E}\left( {{v}_{i}},k \right)-2}{\widehat{{{v}_{l}}{{v}_{l+1}}{{v}_{l+2}}}}
\end{equation}
where $d(u,\ v)$ is the Euclidean distance between nodes $u$ and $v$, and the $\widehat{{{v}_{l}}{{v}_{l+1}}{{v}_{l+2}}}$ is the angle between $v_l$, $v_{l+1}$ and $v_{l+2}$.
The function $\mathbb{E}\left( {{v}_{i}},k \right)$ denotes index $m$ such that node $v_m$ is the k-th nearest neighbor of $v_i$ in Euclidean space.
To compare the difference between two graphs, a complete bipartite graph ${{\mathcal{G}}_{ij}}$ with two node sets $u_i\in G_i$ and $v_j\in G_j$ is constructed.
The weight $w_{ij}$ of each edge in the graph is the Euclidean distance between the node vectors $\boldsymbol{A}_{G_i}$ and $\boldsymbol{A}_{G_j}$.
Next, the Hungarian algorithm is applied to ${{\mathcal{G}}_{ij}}$ to obtain the minimum weighted matching cost, which serves as the geometric discrepancy $\beta$ and the nodes matching matrices $\mathbf{M}_\text{node}$ between $G_i$ and $G_j$:
\begin{equation}
\beta = \min_{\mathbf{M}_\text{node}} \sum_{u\in G_i}\sum_{v\in G_j} w_{uv}(\mathbf{M}_\text{node})_{uv}
\end{equation}
Finally, we use both the area discrepancy $\eta$ and the geometric shape difference $\beta$ to jointly describe the local geometric correlation $\psi_{ij}$ between $G_i$ and $G_j$:
\begin{equation}
{{\psi }_{ij}}=\left( 1-\left| \frac{{{\eta }_{i}}-{{\eta }_{j}}}{\max \left( {{\eta }_{i}},{{\eta }_{j}} \right)} \right| \right) \cdot \exp \left( -z\cdot \beta \right)
\label{eq.geo}
\end{equation}
Here, $z$ is a scaling factor for shape differences ($z>0$), adjusted to align the scale of both types of differences.
The area discrepancy term is built upon $\eta_i$ and $\eta_j$, where $\eta_i$ ($\eta_j$) denotes the area of the closed region enclosed by the vertex sequence of polygon $G_i$ ($G_j$) in the image pixel coordinate system.\\
\subsubsection{Texture similarity correlation}
DSF (Deep-Spectral Factor) $\chi$ is a metric measuring the number of homologous points established between two candidate polygons by a pre-trained deep feature matcher (e.g., LoFTR). It is used in the Local Matcher to determine whether the local geometric constraints for the current polygon pair are sufficient, thereby deciding which correlation computation branch to employ. When the DSF is insufficient for local correction, the texture features are computed directly.
For texture features, to be able to effectively deal with images with large radiometric differences, we use normalized correlation coefficient templates to match similar texture features, as it has radiometric linear invariance and can cope with images with large radiometric differences.
We compute the correlation coefficient $R$ between the image patch $I$ and target patch $T$ within the bounds of the source and candidate polygon.
Finally, we combine local geometric correlation $\psi$ and texture similarity correlation $R$ to calculate the matching cost $\zeta$ between the source polygon and candidate polygons:
\begin{equation}
    R(x,y)=\frac{\sum_{{x}'=0}^{w-1}\sum_{{y}'=0}^{h-1}\mathbb{M}(T({x}',{y}'))\mathbb{M}(I(x+{x}',y+{y}'))}{\sqrt{\left(\sum_{{x}'=0}^{w-1}\sum_{{y}'=0}^{h-1}\mathbb{M}(T({x}',{y}'))^2\right)\left(\sum_{{x}'=0}^{w-1}\sum_{{y}'=0}^{h-1}\mathbb{M}(I(x+{x}',y+{y}'))^2\right)}}
    \label{eq.R}
\end{equation}
Here, $\mathbb{M}$ denotes mean centering, $I$ and $T$ are the image patch and target patch respectively.
$w$ and $h$ are respectively the width and height of the image.
In regions with fewer corresponding points, texture features can still be leveraged to increase the number of matches.
Our experiments later demonstrate that jointly considering geometric and texture features effectively improves both matching accuracy and match quantity.\\
\subsubsection{Local-joint geometry and multi-feature matching}
Local-joint geometry and multi-feature matching strategy (LoJoGM) solves the issue of local topological inconsistencies and abrupt disparity changes in Fig.~\ref{rw:1} and Fig.~\ref{method_local}.
Unlike methods that rely on isolated, greedy correspondence checks, our LoJoGM reframes the polygon matching task as a local-holistic bipartite graph optimization problem, i.e., within each local search region of BPM (Sec.~\ref{sec:BPM}), the source polygons obtain optimal and reliable matching with candidate target polygons, which ultimately leads to a holistic optimal correspondence.
It effectively bypasses the impact of topological inconsistencies and disparity mutations, and improves the matching reliability by confining them to a consistent spatial context because it does not rely on global spatial relationships.
The proposed LoJoGM provides a robust solution to the challenges caused by local disparities and topological variations.

Specifically, we convert the multi-matching relationship between source $\mathcal{A}=\{G_i\}_{i=1}^m$ and target polygons $\mathcal{B}=\{G_j^\prime\}_{j=1}^n$ into a bipartite graph $\mathcal{M}=\left(\mathcal{A},\mathcal{B},\mathcal{E}\right)$, where edges $\mathcal{E}_{ij}$ in the bipartite graph are weighted by the matching cost $\zeta_{ij}$ in Eq.~\ref{eq.cost}.
The cost encapsulates the joint probability of correspondence derived from the geometric discrepancy $\psi_{ij}$ or the texture similarity correlation $R_{ij}$ in Eq.~\ref{eq.geo} or Eq.~\ref{eq.R}, respectively.
To ensure robustness against spectral variations and shape deformations, the cost function is formulated as:
\begin{equation}
    {\zeta _{ij}} = \left\{ {\begin{array}{*{20}{c}}
{\frac{1}{{{\psi _{ij}} \cdot \ln (\chi + e)+ \epsilon}}}&{\chi \ge \gamma}\\
{\frac{1}{{{\psi_{ij}\cdot R_{ij}} +\epsilon }}}&{\chi < \gamma }
\end{array}} \right.
\label{eq.cost}
\end{equation}
Here, $\gamma$ represents the threshold about DSF $\chi$, $\epsilon$ is a small positive constant.
The core innovation of LoJoGM lies in resolving the linear assignment problem to enforce topological coherence.
Instead of independent decision-making, we seek a binary assignment matrix $ \mathbf M=\left[x_{ij}\right]$ that minimizes the total structural energy of the local neighborhood:

\begin{equation}
{\min _\mathbf M}\sum\limits_{i = 1}^{|{\mathcal A}|} {\sum\limits_{j = 1}^{|{\mathcal B}|} {{\zeta _{ij}}} } {x_{ij}},{\text{where}}\quad \sum\limits_j {{x_{ij}}} \le 1,\quad \sum\limits_i {{x_{ij}}} \le 1,\quad {x_{ij}} \in \{ 0,1\}
\end{equation}

This does not enforce a complete assignment for all elements in $\mathcal{A}$, and each polygon $\mathcal{B}_j$ can be matched to at most one polygon in $\mathcal{A}$.
It is essential to clarify that this is an adaptation mechanism that seeks the holistic optimal assignment among local matching candidates, rather than a filtering mechanism. A filtering scheme evaluates each candidate independently against a fixed threshold and commits to each decision irreversibly. In contrast, through its augmenting-path mechanism (i.e., regret mechanism), LHBGO jointly optimizes all candidates within the subgraph and can revoke a tentative assignment whenever a reassignment lowers the overall cost, thereby resolving the conflict ambiguity intrinsic to topological inconsistency rather than merely discarding it. Notably, this assignment relies solely on the holistic consistency of local geometric and multi-feature evidence and does not impose any conventional disparity constraint; consequently, it is inherently insensitive to the abrupt disparity changes and topological inconsistencies that arise when the viewpoint changes.
Crucially, local matching is dedicated to re-matching the polygons that failed in global matching, recovering additional reliable correspondences and partially overcoming topological conflicts and abrupt disparity changes that the global coplanarity assumption cannot accommodate.
Consequently, the method achieves reliable matching by confining solutions to a geometrically consistent manifold, robustly handling the many-to-many ambiguities inherent in stereo vision.

\subsection{Downstream task pipelines}
\label{sec:downstream_method}

\subsubsection{Pose estimation from polygon correspondences}
\label{sec:pose_pipeline}
The matched polygon pairs are used for relative pose estimation, validating the geometric precision of polygon matching.
The core idea is to convert polygon-level correspondences into point-level correspondences, and then estimate camera pose through the standard epipolar geometry framework.
We adopt the geometric area matching protocol \citep{zhangMESAMatchingEverything2024} to enable direct and fair comparison with baselines.
The pipeline proceeds as follows.

First, given a set of matched polygon pairs $\{(P_i, P_j^\prime)\}$, two complementary sets of point correspondences are extracted.
The global set $\mathcal{M}_1$ consists of feature point correspondences obtained by a detector-free matcher (e.g., LoFTR) across the full image pair.
The local set $\mathcal{M}_2$ consists of correspondences extracted within the bounding boxes of each matched polygon pair, providing region-specific densification.

Second, to eliminate matching redundancy, $\mathcal{M}_1$ is spatially filtered using the matched polygon regions as constraints, retaining only correspondences whose source point lies within a source polygon and whose target point lies within the corresponding target polygon.
This filtering leverages the low-redundancy property of LHBGO, where matched polygon regions are mutually non-overlapping, thereby removing ambiguous correspondences at the source.
The filtered global correspondences $\mathcal{M}_1^f$ are merged with the local correspondences $\mathcal{M}_2$ to form the unified set $\mathcal{C} = \mathcal{M}_1^f \cup \mathcal{M}_2$.

Third, a geometric consistency verification based on Sampson distance is applied to reject outlier regions before final pose estimation.
The essential matrix $\mathbf{E}$ is then estimated from $\mathcal{C}$ using the five-point algorithm within MAGSAC \citep{barathMAGSACFastReliable2020a}, which marginalizes over multiple noise scales to perform implicit weighted least-squares estimation without requiring a predefined inlier threshold.
The epipolar constraint is formulated as:
\begin{equation}
    {\mathbf{p}_R}^\top \mathbf{E}\, \mathbf{p}_L = 0
\end{equation}
where $\mathbf{p}_L$ and $\mathbf{p}_R$ are normalized image coordinates of matched points.
The essential matrix is decomposed into the relative rotation $\mathbf{R}$ and translation direction $\boldsymbol{t}$:
\begin{equation}
    \mathbf{E} = [\boldsymbol{t}]_\times \mathbf{R}
\end{equation}
where $[\boldsymbol{t}]_\times$ denotes the skew-symmetric matrix of $\boldsymbol{t}$.
The unique physically feasible solution is determined by the cheirality constraint.

\subsubsection{Polygon cloud generation}
\label{sec:polycloud_pipeline}
\textbf{Definition.} A polygon cloud is formally defined as a set of closed 2D polygonal meshes that conform to the true 3D surfaces of objects. Each polygon element is a surface-conforming closed region that follows the curvature of the underlying object surface. The boundary connecting adjacent polygon vertices is not a simple straight line in 3D space, but a geodesic-like path that traces along the actual object surface, analogous to the concept of geodesics in geosciences. This strict surface-conforming definition distinguishes polygon cloud from planar polygon representations and naturally motivates the algorithmic design that follows.

The optimal polygon surface generation algorithm lifts matched 2D polygons into 3D model space as closed surface-conforming meshes through polygon diffusion approximation and polygon edges of projection surface.
The key principle is to project polygon vertices and boundaries directly onto the triangulated mesh surface via ray casting and surface tracing, ensuring all projected points lie strictly on the reconstructed object surface rather than being interpolated from discrete depth maps.

Given multi-view imagery $\{I_i\}_{i=1}^{N}$, a dense point cloud is first obtained through multi-view stereo reconstruction, followed by surface reconstruction to generate a triangulated mesh model $\mathcal{M} = (\mathbf{V}, \mathbf{F})$.
The camera intrinsic matrices $\mathbf{K}_i$ and extrinsic parameters $[\mathbf{R}_i \mid \boldsymbol{t}_i]$ are obtained simultaneously.
A BVH (Bounding Volume Hierarchy) acceleration structure is built over $\mathcal{M}$ to support efficient ray-mesh intersection.

The optimal polygon surface generation then proceeds as follows.

\begin{figure}[h!t]
    \centering
    \includegraphics[width=0.5\linewidth]{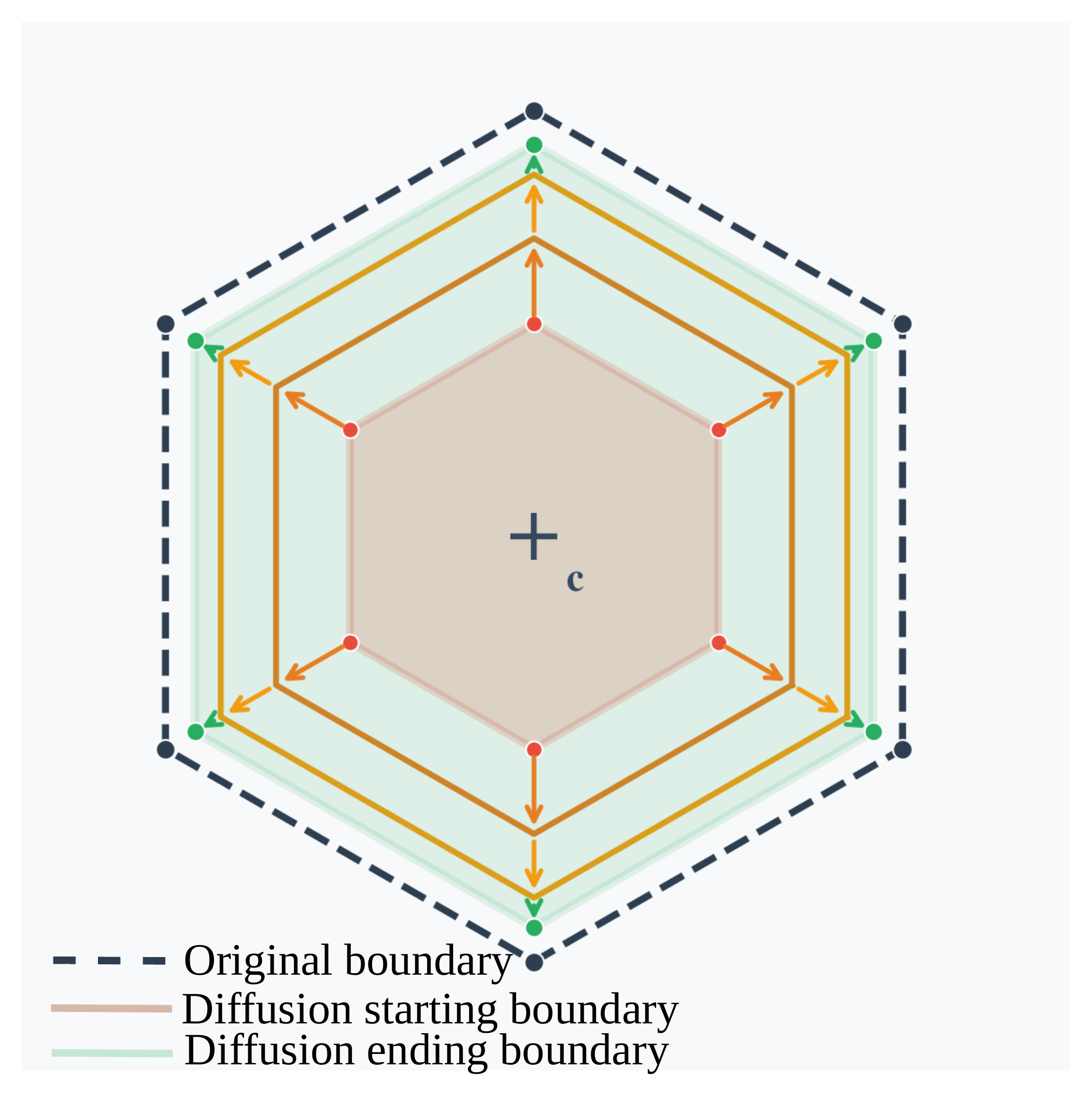}
    \caption{Illustration of the adaptive bisection search for polygon boundary refinement. (a) In 2D image space, a seed polygon is initialized at scale $s_{\min}$ from the centroid and progressively expanded toward the original boundary via binary search. (b) At each iteration, the scaled polygon is projected onto the reconstructed mesh via ray casting; if 3D quality constraints are satisfied, the lower bound is raised ($lo \leftarrow mid$), otherwise the upper bound is reduced ($hi \leftarrow mid$), until convergence yields the maximum valid scale $s^*$.}
    \label{fig:seed_polygon_expansion}
\end{figure}

\textbf{(1) Polygon diffusion approximation.}
The root cause of unreliable boundary projection lies in 3D space: the reconstructed 3D object usually exhibits jagged, low-accuracy point clouds and meshes along its edges. Consequently, naively pairing each 2D boundary point with its corresponding 3D edge point easily lets the projection ray slip past the foreground mesh and land on distant background surfaces, producing severe geometric distortion.
Hole-filling is first applied to $\mathcal{M}$ to maximize surface completeness, because any missing region would cause ray casting to miss the surface, leaving 2D polygon vertices without valid 3D correspondences.
Subsequently, a polygon diffusion approximation algorithm iteratively expands outward from the polygon centroid toward the boundary vertices, using binary search to find the maximum valid projection extent (Fig.~\ref{fig:seed_polygon_expansion}).
Starting from a conservative scale $s = s_{\min}$, the boundary is progressively diffused outward by testing increasing scale factors $s \in [s_{\min}, 1.0]$.
At each trial scale, the scaled polygon vertices are projected onto the mesh via ray casting, and geometric quality constraints are evaluated: (1) inter-vertex edge length within a physical plausibility threshold, ensuring no adjacent vertices land on disparate surfaces; (2) local smoothness measured by angular deviation between adjacent vertex normals, rejecting sharp folds indicative of boundary-crossing; (3) spatial bounding box within scene-appropriate scale.
Binary search over $s$ converges to the largest valid scale that preserves boundary fidelity while guaranteeing all projected vertices reside on the correct foreground surface.
This design maximizes boundary coverage, which directly determines the geometric completeness of the final polygon cloud.

\begin{figure}[h!t]
    \centering
    \includegraphics[width=\linewidth]{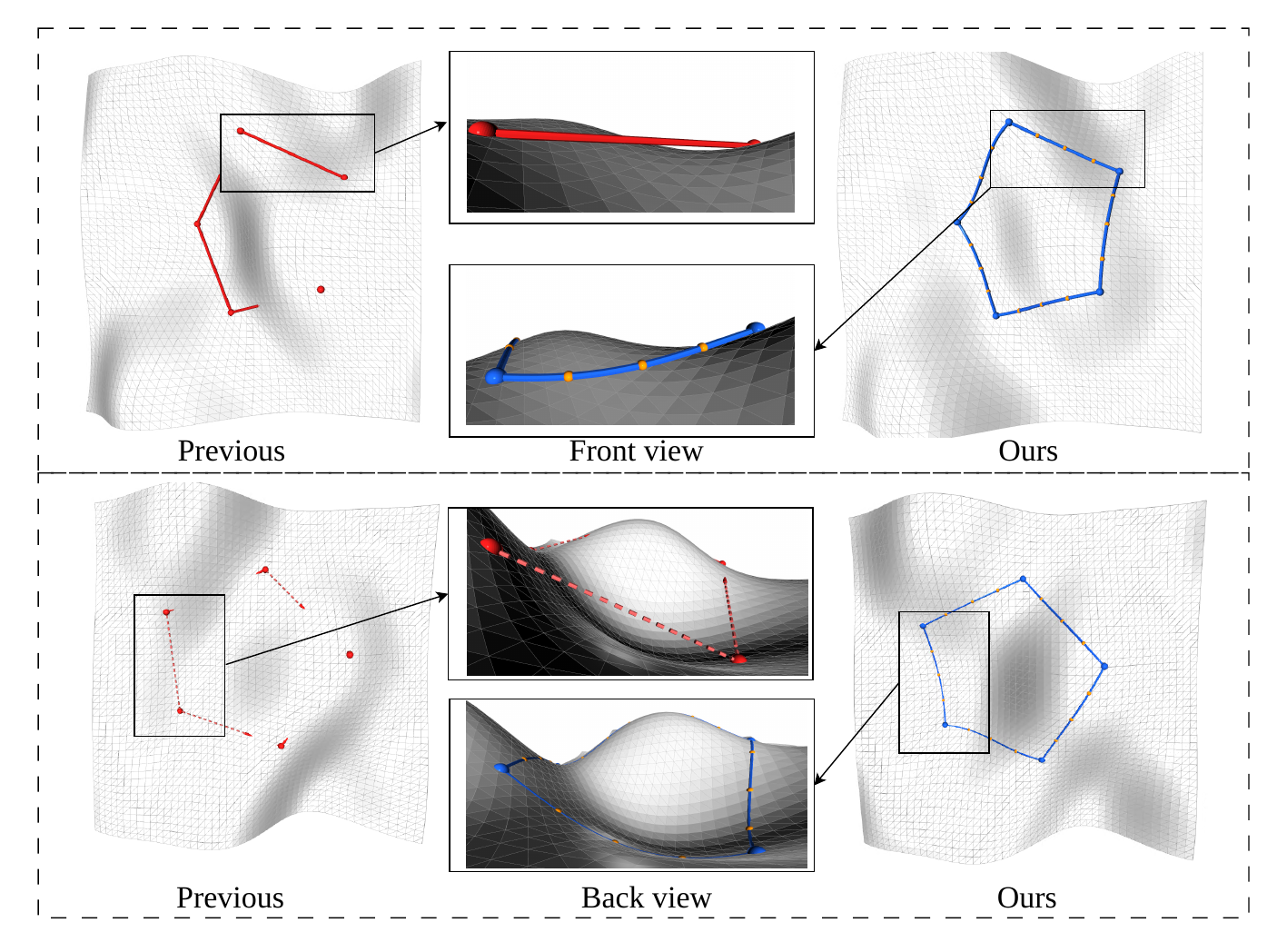}
    \caption{Comparison of naive vertex projection (left) and the proposed polygon edges of projection surface strategy (right). In naive projection, polygon edges directly connecting sparse 3D vertices penetrate the mesh surface (dashed lines indicate occluded segments). With boundary densification and polygon edges of projection surface, cross-points are inserted at mesh edge intersections (orange), producing a closed contour that strictly conforms to the triangulated surface.}
    \label{fig:surface_line_tracing}
\end{figure}

\textbf{(2) Polygon edges of projection surface.}
After projecting polygon vertices onto the mesh, adjacent 3D anchor points $\mathbf{X}_l$ and $\mathbf{X}_{l+1}$ must be connected along the mesh surface.
Our key insight is that a surface-conforming edge can be obtained directly by densification and projection, without solving any continuous variational problem. Specifically, the 2D boundary segment between two adjacent vertices is first densified into a sequence of sample points at a fixed pixel interval $\delta$, $\{(u_k,v_k)\}_{k=0}^{K}$. Each sample is then projected onto the mesh surface $\mathcal{M}$ via ray casting (Eq.~\ref{eq:raydir}), yielding ordered 3D points $\{\mathbf{X}_k\}_{k=0}^{K}$ that all lie strictly on the surface. The polygon edges of projection surface are then defined as the polyline connecting these projected points in order:
\begin{equation}
    \boldsymbol{\gamma}^{*} = \big(\mathbf{X}_0, \mathbf{X}_1, \ldots, \mathbf{X}_K\big),
    \quad \mathbf{X}_k = \Pi_{\mathcal{M}}\!\big(u_k, v_k\big),\ \ \mathbf{X}_k \in \mathcal{M},
\end{equation}
where $\Pi_{\mathcal{M}}(\cdot)$ denotes the ray-casting projection onto $\mathcal{M}$. Geometrically, this projected edge lies on the intersection between the mesh surface $\mathcal{M}$ and the (approximately orthogonal) plane spanned by the original 3D object-space edge and the surface, so it is an optimal surface-conforming edge that stays as close as possible to the straight object-space edge while remaining on the surface, and in practice it also coincides with the shortest path along the surface in the common case of locally smooth regions. In this way, the densify-then-project scheme directly produces a surface-conforming edge in closed form, avoiding any explicit variational optimization.
As shown in Fig. \ref{fig:surface_line_tracing}, a straight-line connection in 3D space would penetrate the mesh interior, violating the surface-conforming requirement.
Polygon edges of projection surface address this by traversing the mesh triangle by triangle between adjacent anchors:
(1) within the current triangle's local tangent plane, the projected direction vector toward the target anchor $\mathbf{X}_{l+1}$ is computed;
(2) the first mesh edge that the direction vector crosses is identified, and the exact intersection point on that edge is calculated;
(3) the crossing point $\mathbf{P}_{\text{cross}}$ is inserted as an intermediate contour node (obtained by linear interpolation between edge endpoints, guaranteeing it lies strictly on the mesh surface), and the traversal transitions to the adjacent triangle sharing that edge;
(4) the tracing terminates when $\mathbf{X}_{l+1}$ lies within the current triangle (verified by barycentric coordinate test).
This expands each adjacent anchor pair into $\mathbf{X}_l \to \mathbf{P}_1 \to \mathbf{P}_2 \to \cdots \to \mathbf{X}_{l+1}$, where all intermediate points reside on mesh edges and connected segments travel strictly along triangle surfaces, fundamentally preventing mesh penetration.
This approach generates geometrically exact paths on the continuous mesh surface rather than approximate paths constrained to discrete mesh vertices.
It is worth clarifying that operating on mesh edges does not discard 3D spatial information. Each crossing point $\mathbf{P}_{\text{cross}}$ is a full 3D coordinate obtained by linear interpolation between the two 3D endpoints of the mesh edge it lies on, so the traversal is carried out entirely in 3D model space rather than in a flattened 2D domain. Because every segment is constrained to a triangular face and every bend occurs on a shared edge, the reconstructed boundary, once lifted back to 3D, is guaranteed to run along the actual object surface.

\textbf{(3) Per-point ray casting.}
For each boundary sample point $(u_l, v_l)$ of the polygon (after edge densification at fixed pixel intervals $\delta$), a ray is cast from the camera center $\mathbf{O} = -\mathbf{R}^{\top}\boldsymbol{t}$ along the viewing direction:

\begin{equation}
    \mathbf{d}_l = \mathbf{R}^{\top} \cdot \mathbf{K}^{-1} \begin{pmatrix} u_l \\ v_l \\ 1 \end{pmatrix}
    \label{eq:raydir}
\end{equation}
The ray $\mathbf{r}(t) = \mathbf{O} + t \cdot \hat{\mathbf{d}}_l$ is intersected with $\mathcal{M}$ using the BVH structure.
The first hit point $\mathbf{X}_l = \mathbf{O} + t_{\text{hit}} \cdot \hat{\mathbf{d}}_l$ lies on a specific triangular face, inherently guaranteeing surface conformity without depth-map quantization noise.

\textbf{(4) Outlier rejection.}
A multi-criteria quality check is applied to the projected 3D contour: if the maximum inter-vertex distance exceeds $\tau$ times the median distance, the polygon is considered to straddle a depth discontinuity and is rejected.
Local fold detection further rejects polygons with excessive angular deviation between adjacent triplets.
The accepted polygon boundaries are rendered as bold 3D contour lines with interiors densely sampled from the mesh surface.
Each polygon retains its instance-level identity from the matching stage, producing a polygon cloud where each element is a closed surface-conforming mesh patch with semantic labels.

\subsection{Metric for polygon matching}
\subsubsection{Ground truth generation of polygon matches}
Unlike point/line correspondence-level evaluation, polygon matching tasks need to consider discrete metrics (i.e., precision and recall) rather than global errors due to geometric properties.
Because the discrete metrics are evaluated by the ground truth (GT) of polygon matching, we design a scheme to generate the GT of polygon matching.
We project the geometric center $\boldsymbol{x}_i=\left(p,q,1\right)^\top$ of polygon $G_i$ to ${\boldsymbol{x}^\prime}_i$ in $G_j$ onto the target image:
\begin{equation}
    {{\boldsymbol{x}'}_{i}} = {{\mathbf{P}}_{1}}\cdot {{\left( \mathbf{T}_{\text{R}}^{-1}\cdot {{\mathbf{T}}_{\text{L}}}\cdot d\cdot {{\left[ \begin{matrix}
   \mathbf{P}_{0}^{-1}{{x}_{i}} & 1  \\
\end{matrix} \right]}^{\top}} \right)}_{\text{proj}}}
\end{equation}
Here, $\mathbf{P}_0$ and $\mathbf{P}_1$ are the intrinsic matrices of the left and right cameras, respectively.
$\mathbf{T}_R$ and $\mathbf{T}_L$ are the transformation matrices of the right and left cameras, respectively.
$d$ denotes the depth value of the image.
The projection operator ${{\cdot }_{\text{proj}}}$ converts normalized coordinates to pixel coordinates by dividing them by the depth dimension.
Then, we leverage a bidirectional consistency check to refine the GT of matching relationships $M_{ij}$:
\begin{equation}
    {{\mathbf{M}}_{ij}}=\mathbb{I}(|{{x}_{i}}-{{{x}'}_{i}}|{{|}_{2}}\le \lambda )
    \label{eq.gt}
\end{equation}
Here, the indicator function $\mathbb{I}(\cdot)$ returns 1 if the specified condition is satisfied and 0 otherwise.
The parameter $\lambda$ is an empirical threshold.
Discrete metrics in polygon matching can be evaluated through the GT of homologous image points and the geometric properties of polygons, which are often unmeasurable in point or line matching tasks.\\

\subsubsection{Matching area score}
To quantitatively evaluate the geometric precision of our matching results, we define the matching area score (MAS).
This metric is designed to be robust to topological inconsistencies in segmentation by comparing the total matched area.

Let the sets of predicted matched polygons in the left and right images be $P_{L}^{\text{pred}}$ and $P_{R}^{\text{pred}}$, and the corresponding ground truth sets be $P_{L}^{\text{gt}}$ and $P_{R}^{\text{gt}}$.
We first compute a single unified region for each set via $\mathcal{U}(P)$.
The MAS is then the mean of the IoU calculated as follows:
\begin{equation}
    \text{MAS} = \frac{1}{2} \Big( \text{IoU}(\mathcal{U}(P_{L}^{\text{pred}}), \mathcal{U}(P_{L}^{\text{gt}}))
    + \text{IoU}(\mathcal{U}(P_{R}^{\text{pred}}), \mathcal{U}(P_{R}^{\text{gt}})) \Big)
\end{equation}

The primary advantage of MAS is its robustness to segmentation differences.
By aggregating all matched polygons into a single region before computing IoU, the MAS score correctly evaluates the total spatial overlap, making it insensitive to one-to-many or many-to-one correspondences.
This ensures algorithms are rewarded for accurately identifying the overall matched region, rather than being penalized for minor differences in polygonization.
Furthermore, its formulation provides an unbiased assessment, independent of which image is considered the source.

\subsubsection{Area coverage ratio}
To provide a more accurate and meaningful evaluation of matching performance, we propose the Area Coverage Ratio (ACR).
This metric is designed as a refinement of the Area of Overlap Ratio (AOR) found in benchmarks like MESA \citep{zhangMESAMatchingEverything2024}.
A primary limitation of AOR is that it measures the matched area against the total image area, which can produce misleadingly high scores for methods that result in significant over-coverage.
An algorithm could incorrectly match large, irrelevant regions and still be rewarded.

The ACR addresses this flaw by focusing the evaluation specifically on the objects of interest.
It is calculated not against the entire image, but as the ratio of the total area of all predicted matched polygons to the total area of all ground truth polygons available for matching.
This ensures the metric directly reflects the algorithm's capability to identify semantically significant regions, providing a more robust and intuitive measure of performance that penalizes both missed polygons and irrelevant matches.
The formula is given by:

\begin{equation}
    \text{ACR} = \frac{\sum \text{Area}(\text{matched})}{\sum \text{Area}(\text{GT})}
\end{equation}

\section{Experiments}
\label{sec:exp}
\subsection{Experimental setup}
\subsubsection{Datasets and baselines}
\method was evaluated on five challenging public datasets (SceneFlow \citep{mayerLargeDatasetTrain2016}, KITTI \citep{Menze2018JPRS}, ScanNet \citep{daiScanNetRichlyAnnotated3D2017}, DTU \citep{jensenLargeScaleMultiview2014} and ISPRS large-format images \citep{rottensteinerIsprsTestProject2013}). This paper involves three aspects of experiments with corresponding baselines:

(1) \textit{Polygon matching experiments.} Since no existing polygon matching methods from stereo imagery have been available for direct comparison, we selected related SoTA approaches and adapted their outputs. MESA \citep{zhangMESAMatchingEverything2024} is a region-based matching approach, SGAM \citep{zhangSearchingAreaPoint2024a} and MASA \citep{liMatchingAnythingSegmenting2024} are other related baselines. Additionally, we incorporate two classical shape matching methods, Shape Context \citep{belongieShapeMatchingObject2002} and IDSC \citep{lingShapeClassificationUsing2007}, as baselines representing the shape retrieval paradigm (marked with $\dagger$ in Table~\ref{exp_tabel:1}).

(2) \textit{Pose estimation experiments.} We compare \method against representative point-level baselines (SP\citep{detoneSuperPointSelfsupervisedInterest2018}+SGMNet\citep{chenLearningMatchFeatures2021} and SiLK\citep{gleizeSiLKSimpleLearned2023}), the dense matcher LoFTR\citep{sunLoFTRDetectorfreeLocal2021}, and area-level matchers (SEEM\citep{NEURIPS2023_3ef61f7e}+SGAM+LoFTR and MESA+LoFTR) on the ScanNet1500 benchmark.

(3) \textit{Polygon cloud reconstruction.} As the first work to propose and construct polygon cloud, no existing baselines are available for direct comparison. We provide qualitative evaluation on the DTU dataset to demonstrate the feasibility of this innovative 3D representation.

We employ MAS for matching completeness, under three thresholds (80/50/40, \%), and precision as discrete metrics. For the downstream pose estimation task, we report the area under the cumulative error curve (AUC) of the pose error at thresholds of $5^\circ$, $10^\circ$, and $20^\circ$, following the standard ScanNet1500 evaluation protocol, where higher AUC values indicate more accurate pose estimation.

\subsubsection{Comparative experimental setup}
To comprehensively evaluate the diverse capabilities of \method, we selected specific datasets and metrics tailored to the distinct objectives of each experiment:
\begin{enumerate}[(1)]
    \item SceneFlow, KITTI and ScanNet for quantitative evaluation (Table~\ref{exp_tabel:1} and Table~\ref{tab:discrete_exp}).
    We utilized the SceneFlow, KITTI and ScanNet datasets for our primary comparative evaluation because they provide complete, high-quality depth maps and intrinsic parameters.
    The availability of precise depth information is a prerequisite for our proposed GT generation pipeline (Eq.~\ref{eq.gt}), which projects polygons between views to rigorously calculate discrete metrics. Consequently, we employed MAS on these datasets to strictly quantify the geometric accuracy and completeness of the matches.
    \item ISPRS for scalability and efficiency (Table~\ref{exp_tabel:5}).
    To validate the effectiveness of the BPM on large-format images, we employed the ISPRS dataset.
    Unlike standard benchmarks, ISPRS consists of large-format aerial imagery with resolutions significantly higher than typical computer vision datasets.
    This makes it the ideal testbed for assessing computational efficiency and the ability to handle massive scale variations.
    For these experiments, we focused on ACR and Runtime to demonstrate how our pyramid strategy maintains high coverage while significantly reducing computational costs compared to full-resolution baselines.
    \item ScanNet for camera pose estimation (Table~\ref{tab:pose_estimation}).
    The ScanNet dataset was adopted for pose estimation because it provides ground truth camera trajectories and presents challenging conditions with poor texture and large viewpoint variation, which is consistent with the evaluation protocol of MESA \citep{zhangMESAMatchingEverything2024}. Due to the difficulty of some image pairs, matching may fail.
    This experiment validates the geometric quality of polygon matching outputs for photogrammetric downstream applications.
    \item DTU for polygon cloud reconstruction (Fig. \ref{fig:polygon_cloud}).
    The DTU dataset provides high-quality multi-view imagery with rich and varied surface geometry, making it suitable for qualitative evaluation of 3D polygon cloud generation.
    This experiment demonstrates that polygon matching results can be directly projected into structured 3D representations with surface-conforming boundaries, bridging 2D matching with 3D scene understanding.

\end{enumerate}
\subsubsection{Experimental environment}
All experiments were conducted on one NVIDIA L40.
After our tests, the minimum hardware requirement is RTX2060, and the video memory requirement does not exceed 6GB.
We summarize the key hyperparameters used in our experiments as follows, and all hyperparameters used in our method are empirically determined based on tests.
All images are kept at their original resolutions: $1296\times968$ for ScanNet, $960\times540$ for SceneFlow, and $1242\times375$ for KITTI.
The search window size for the source polygon is $15\times15$, and the initial search range for the target polygon is $50\times50$.
In Algorithm \ref{alg:1}, the pyramid downsampling factor is set to 3, with the width or height of the top-level image limited to 200 pixels.
The local window size $\tau$ is set to 25.
In LoJoGM, the similarity threshold $\iota$ is set to 5.
The value of $\gamma$ is 8.
The confidence level $\lambda$ is 0.1.
Scaling factor $z$ is 5.

\subsection{Quantitative experiment}
\subsubsection{Polygon matching}
As presented in Table~\ref{exp_tabel:1}, we conducted a comprehensive evaluation of \method against state-of-the-art methods (MESA, SGAM, and MASA) across the SceneFlow, KITTI, and ScanNet datasets.
The results highlight the robustness and efficiency of our proposed framework under varying scene complexities.

The classical shape matching methods (Shape Context and IDSC, marked with $\dagger$) consistently produce the highest number of matched pairs across all datasets (e.g., 60.37 on SceneFlow, 55.37 on KITTI) because they perform exhaustive pairwise shape descriptor comparison without any search space reduction.
Consequently, this exhaustive strategy incurs prohibitively long computation times (e.g., IDSC requires 902s on SceneFlow) and yields substantially lower MAS than our method, as purely geometric descriptors are insufficient to disambiguate polygons with similar shapes in stereo imagery.

On the synthetic SceneFlow and outdoor KITTI datasets, \method combined with SuperPoint and LightGlue consistently achieves the best performance across all MAS thresholds. It achieves an MAS$_{40}$ of 68.60\% and 58.20\%, respectively, surpassing the strongest area-level baseline SGAM by absolute margins of 28.29\% and 31.80\%. Notably, on the KITTI dataset, our method improves MAS by over 31\% compared to SGAM (26.40\%), while maintaining a high number of correct matches (30.28).
In terms of efficiency, the SuperPoint and LightGlue configuration achieves real-time-comparable processing speed, processing KITTI image pairs in just 1.92s, which represents a $53.75\times$ speedup over MESA (103.20s). While MASA achieves high speeds, it produces an insufficient number of valid matches (approx.2 matches per pair), rendering it ineffective for practical polygon matching tasks.

The indoor ScanNet dataset presents significant challenges due to low texture, occlusion, and complex geometries.
In this domain, the \method variant equipped with LoFTR achieves the best MAS$_{40}$ of 21.25\%, outperforming all baselines including SGAM (16.94\%) and MESA (13.52\%). This indicates that LoFTR's dense matching capability provides stronger geometric constraints in texture-poor indoor environments.
Across all three datasets, \method demonstrates consistent superiority in MAS, validating the effectiveness of combining polygon-level structural constraints with learned features for zero-shot polygon matching.

It is notable that MAS values remain nearly constant across different thresholds (40/50/80) for all methods.
This indicates that the matched polygon pairs exhibit high geometric discriminability: once a polygon pair satisfies the matching criterion at a stricter threshold, it almost invariably satisfies looser thresholds as well.
In other words, the polygon matching results produced by all methods exhibit a clear binary distinction between correct matches and mismatches, with few borderline cases that would be sensitive to threshold selection.

\begin{table}[h!tb]
\centering
\caption{Evaluation on SceneFlow, KITTI, and ScanNet datasets.
\textbf{Bold} indicates the best result and \underline{underline} indicates the second best within each dataset.
$\dagger$ denotes classical shape matching methods.}
\resizebox{\linewidth}{!}{
\begin{tabular}{llccccc}
\toprule
\multirow{2}{*}{Dataset} & \multirow{2}{*}{Method}
& \multicolumn{3}{c}{Matching area score (\%)$\uparrow$}
& \multirow{2}{*}{Matches$\uparrow$}
& \multirow{2}{*}{Time(s)$\downarrow$} \\
\cmidrule(lr){3-5}
 & & MAS$_{40}$ & MAS$_{50}$ & MAS$_{80}$ & & \\
\midrule

\multirow{9}{*}{SceneFlow}
  & Shape Context$^\dagger$
    & 43.15 & 43.26 & 43.66 & 60.37 & 247.35 \\
  & IDSC$^\dagger$
    & 43.24 & 43.35 & 43.75 & 60.37 & 902.14 \\
  \cmidrule(lr){2-7}
  & \multicolumn{6}{l}{\textit{Area-assisted / Tracking-based}} \\
  & MESA & 36.23 & 36.20 & 35.96 & 12.05 & 47.52 \\
  & SGAM & 40.31 & 40.29 & 40.13 & 16.00 & 11.08 \\
  & MASA & 29.19 & 30.37 & 30.06 & 3.04 & 2.18 \\
  \cmidrule(lr){2-7}
  & \textbf{Z(PM)$^2$ {\scriptsize +SuperPoint+LightGlue} (Ours)}
    & \textbf{68.60} & \textbf{68.57} & \textbf{68.51} & 27.50 & 3.27 \\
  & Z(PM)$^2$ {\small +LoFTR}
    & \underline{44.95} & \underline{44.91} & \underline{44.74} & 20.59 & 16.94 \\

\midrule

\multirow{9}{*}{KITTI}
  & Shape Context$^\dagger$
    & 38.25 & 38.26 & 38.62 & 55.37 & 215.35 \\
  & IDSC$^\dagger$
    & 38.31 & 38.37 & 38.71 & 55.37 & 792.14 \\
  \cmidrule(lr){2-7}
  & \multicolumn{6}{l}{\textit{Area-assisted / Tracking-based}} \\
  & MESA & 22.15 & 22.14 & 21.85 & 24.26 & 103.20 \\
  & SGAM & 26.40 & 26.35 & 26.17 & 19.14 & 12.83 \\
  & MASA & 18.46 & 18.71 & 18.80 & 2.10 & 1.42 \\
  \cmidrule(lr){2-7}
  & \textbf{Z(PM)$^2$ {\scriptsize +SuperPoint+LightGlue} (Ours)}
    & \textbf{58.20} & \textbf{58.15} & \textbf{58.04} & 30.28 & 1.92 \\
  & Z(PM)$^2$ {\small +LoFTR}
    & \underline{41.26} & \underline{41.20} & \underline{41.00} & 29.55 & 7.77 \\

\midrule

\multirow{9}{*}{ScanNet}
  & Shape Context$^\dagger$
    & 14.52 & 14.91 & 15.70 & 40.96 & 174.63 \\
  & IDSC$^\dagger$
    & 14.66 & 15.06 & 15.85 & 40.96 & 540.98 \\
  \cmidrule(lr){2-7}
  & \multicolumn{6}{l}{\textit{Area-assisted / Tracking-based}} \\
  & MESA & 13.52 & 13.50 & 13.63 & 6.22 & 25.41 \\
  & SGAM & 16.94 & 16.92 & 17.00 & 11.23 & 7.08 \\
  & MASA & 14.50 & 14.13 & 14.66 & 2.03 & 1.20 \\
  \cmidrule(lr){2-7}
  & Z(PM)$^2$ {\scriptsize +SuperPoint+LightGlue} (Ours)
    & \underline{17.39} & \underline{17.41} & \underline{17.59} & 5.01 & 2.48 \\
  & \textbf{Z(PM)$^2$ {\small +LoFTR} (Ours)}
    & \textbf{21.25} & \textbf{21.50} & \textbf{21.05} & 6.32 & 1.42 \\

\bottomrule
\end{tabular}
}
\label{exp_tabel:1}
\end{table}

To further evaluate the fine-grained geometric precision of the matched polygons, we report discrete metrics in Table~\ref{tab:discrete_exp}.
Since competing methods do not output explicit vector-to-vector correspondences suitable for this granular analysis, they are excluded from this comparison.
\method exhibited strong geometric consistency, achieving a Precision of 87.50\% and 82.36\% on SceneFlow and KITTI, respectively.
Even in the challenging ScanNet dataset, the method maintained a Precision of 61.84\% and an F1 score of 0.53, verifying its capability to generate geometrically valid and semantically meaningful polygon matches across diverse domains.

\begin{table}[h]
    \centering
    \caption{Geometric precision metrics and coverage performance. Since the other methods do not perform polygon matching, they cannot be directly compared with \method on discrete metrics.}
    \resizebox{0.6\linewidth}{!}{
        \begin{tabular}{@{}lccccccc@{}}
            \toprule
            Dataset & Recall(\%)↑ & Precision(\%)↑ & F1↑ & ACR(\%)↑ \\ \midrule
            SceneFlow & 72.67 & 87.50 & 0.79 & 89.34 \\
            ScanNet & 46.12 & 61.84 & 0.53 & 83.28 \\
            KITTI & 70.92 & 82.36 & 0.76 & 81.26 \\ \bottomrule
        \end{tabular}
    }

    \label{tab:discrete_exp}
\end{table}

To investigate the dependence of \method on the segmentation model (e.g., SAM), we tested SAMs with varying parameter counts.
Table~\ref{exp_table:sam} shows that the smallest model yielded the fewest segmentations yet achieved higher precision.
The largest model performed best across all metrics, while the medium model offered balanced performance.

\begin{table}[h]
    \centering
    \caption{ Impact of SAM segmentation quality with different parameter sizes on \method matching performance.}
    \begin{tabular}{cccc}
    \hline
       Model & Recall(\%) & Precision(\%) & F1 \\
    \hline
       vit-h & 72.72 & 87.15 & 0.79 \\
       vit-l & 57.11 & 78.40 & 0.66\\
       vit-b & 59.67 & 82.60 & 0.69\\
    \hline
    \end{tabular}
    \label{exp_table:sam}
\end{table}

\subsubsection{Matching in large-format imagery}
To validate the scalability and efficiency of our proposed \method framework in challenging imagery, we conducted experiments on large-format stereo-image pairs.
The ISPRS remote dataset \citep{rottensteinerIsprsTestProject2013} presents significant challenges due to its high resolution between images.
Specifically, we assessed the impact of our bidirectional pyramid matching strategy on both matching accuracy and computational efficiency.
We hypothesized that the pyramid approach would significantly improve computational efficiency by reducing runtime while maintaining acceptable Area Coverage Ratio (ACR).
Table~\ref{exp_tabel:5} presents the quantitative results of our experiments, showcasing the effectiveness of \method in handling large-format images.

\begin{table}[h!]
\centering
\caption{Evaluation of \method on ISPRS dataset under different pyramid levels.
\method(Full) denotes the levels ensuring the top-layer resolution of the constructed image pyramid does not exceed 200 pixels.
The baseline uses a fixed range instead of the bidirectional pyramid.}
\setlength{\tabcolsep}{1mm}{
 \begin{tabular}{@{}lllll@{}}
    \toprule
    Method & Resolution & Time(s)↓ & ACR(\%)↑ & Num↑ \\ \midrule
    \method(Full)                           & 5750$\times$3750  & 10.49              & 21.92    & 17.10 \\
    \method($\mathcal{L}=2$) & 5750$\times$3750  & 10.70              & 21.92    & 17.10 \\
    baseline & 5750$\times$3750 & 15.66 & 21.92 & 17.10 \\\hline
    \method(Full)                           & 11500$\times$7500 & 18.95              & 23.15    & 10.7  \\
    \method($\mathcal{L}=2$) & 11500$\times$7500 & 18.96              & 23.15    & 10.7  \\
    baseline & 11500$\times$7500 & 24.86 & 23.15 & 10.7 \\ \bottomrule
\end{tabular}
}

    \label{exp_tabel:5}
\end{table}

\subsection{Qualitative experiment}
\begin{figure}
    \centering
    \includegraphics[width=\linewidth]{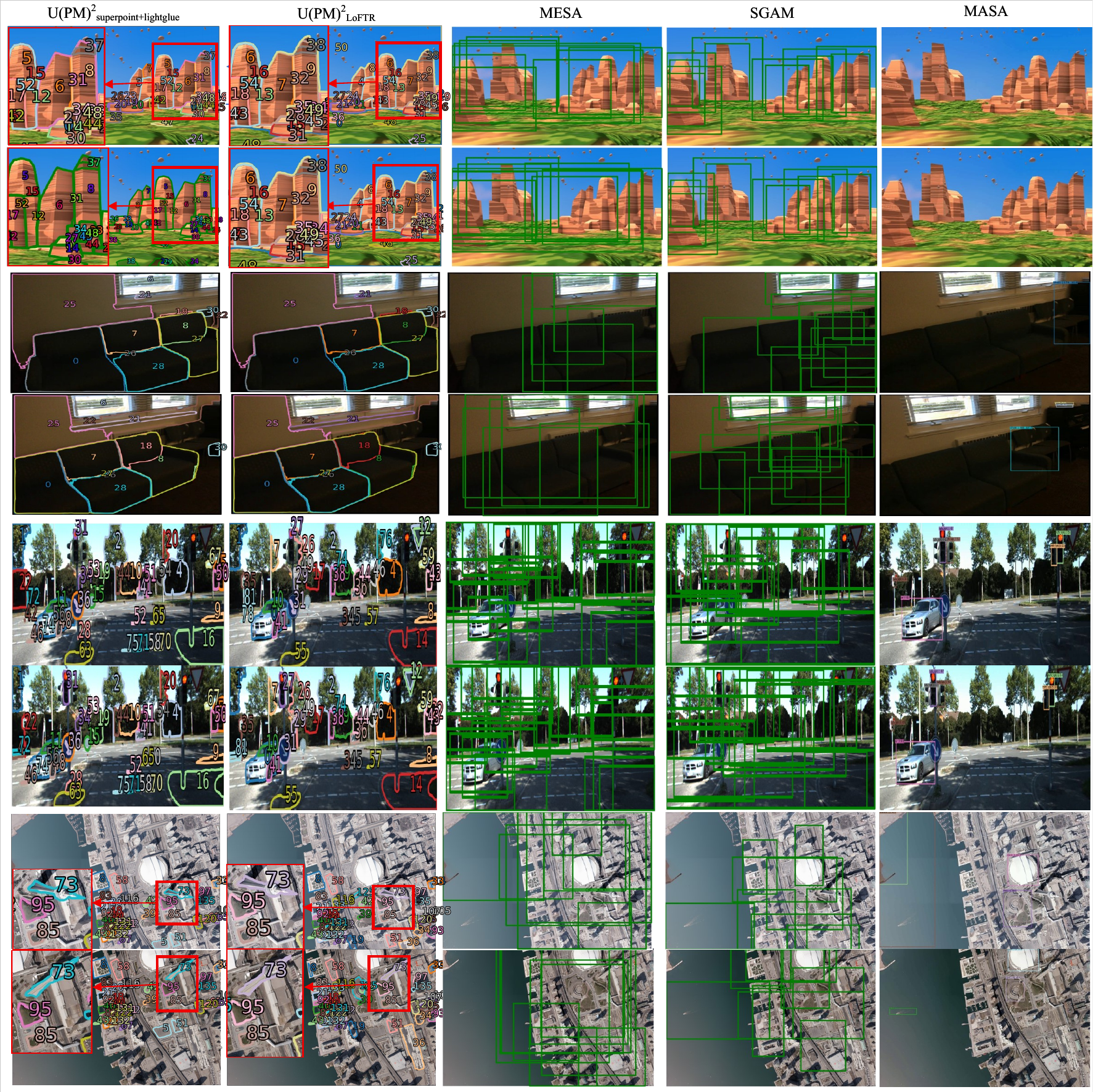}
    \caption{Qualitative results on the four datasets. In the same column of images, the images in the upper and lower rows that are adjacent to each other form a pair of stereo images.}
    \label{fig:Qualitative}
\end{figure}

We present qualitative results on the SceneFlow and ScanNet datasets in Fig.~\ref{fig:Qualitative}, respectively.
Compared to other methods, our method performs better in capturing fine details and structures.
More importantly, our \method can correspond to each nonsemantic patch in stereo image pairs.
\method exhibits consistent performance across different combinations of feature detectors and matchers, as both settings successfully align corresponding polygons in the images.
This robustness stems from its core components (i.e., the bidirectional pyramid matching strategy and LoJoGM) which effectively handle disparity discontinuities and enforce globally optimal matching.
These capabilities are beyond the scope of conventional point-based matchers.
In contrast, competing methods such as MESA and SGAM exhibit significant matching redundancy, resulting in highly overlapping matched regions.
This overlap introduces ambiguity in polygon matching.
MASA produces insufficient matches owing to its training requirements.
To address this challenge, \method introduces a dedicated local matcher, specifically designed to suppress redundant matches and ensure precise correspondence without any training.

\subsection{Ablation study}
To evaluate the effectiveness of each module in \method, we conducted a comprehensive ablation study (Table~\ref{exp_tabel:6}).
After removing pyramid matching and LoJoGM, we observed significant decreases in all metrics, particularly after removing LoJoGM.
The results demonstrate these modules are indispensable for the polygon matching task.

\subsubsection{Bidirectional pyramid matching (BPM)}
The goal of BPM is to provide an adaptive search window.
For the ablation of the pyramid matching, we employed a standard approach to define the search window.
Specifically, for each source polygon $P_i$ in the left image of the stereo pair, a search range $R_i$ is defined in the right image with a radius corresponding to the shortest side of the bounding box $B_i$ around $P_i$.
During global matching, candidate polygons are searched within the range $R_i$.
Most polygons have smaller search ranges $R_i$, so fewer polygons need to be matched during the subsequent local matching stage, leading to faster processing, though this also results in fewer final matches.
Moreover, for more challenging tasks, such as matching small targets in large-format images, the static nature of $R_i$ can lead to more misses due to its smaller scale compared to the overall image size.
\begin{table}[h!]
\centering
\caption{Ablation study on the SceneFlow dataset.}
\begin{tabular}{llll}
\hline
Method & ACR(\%)↑ & Num↑ & Time(s)↓ \\ \hline
\method                          & 89.02 & 42.22 & 3.60 \\ \hline
w/o pyramid & 76.94 & 37.94 & 3.02 \\
w/o LoJoGM & 14.21 & 12.93 & 3.33 \\
w/o LoJoGM \& pyramid & 9.56 & 10.44 & 2.16 \\ \hline
\end{tabular}
\label{exp_tabel:6}
\end{table}\\
\subsubsection{LoJoGM}
LoJoGM achieves fine matches by leveraging local features based on multiple correlations.
For the ablation of this module, we replace the LoJoGM module with a greedy optimization algorithm.
Specifically, each source polygon selects the target polygon with the highest geometric similarity.
Although this may not always lead to a globally optimal solution, it is optimal for each local match.
The final matching performance in Table~\ref{exp_tabel:6}, evaluated by the ACR metric, decreases by approximately 80\% compared to \method.
This demonstrates that relying solely on the geometric features of polygons is inadequate for polygon matching, as these features are not robust to changes in viewpoint and segmentation variations.

\subsection{Downstream tasks}

\subsubsection{Camera pose estimation}
Following the geometric area matching pipeline described in Sec.~\ref{sec:pose_pipeline}, the matched polygon pairs produced by \method serve as spatial constraints for extracting and filtering point correspondences, which are then fed into the essential matrix estimation framework for relative pose recovery.
Experiments were conducted on the ScanNet dataset, which presents challenging indoor conditions with poor texture and large viewpoint variation.
We adopt the standard ScanNet1500 evaluation protocol and compare \method against representative area-level and point-level baselines, with all area-level methods built upon the same LoFTR matcher for a fair comparison.
Table~\ref{tab:pose_estimation} reports the pose estimation accuracy measured by the AUC of the pose error at thresholds of $5^\circ$, $10^\circ$, and $20^\circ$, where higher values indicate better accuracy.
\begin{table}[h!tb]
\caption{Camera pose estimation on ScanNet1500. Results for competing methods are taken from original paper\citep{zhangMESAMatchingEverything2024} when available; otherwise, they are reproduced using the official implementations. All methods are evaluated under the same dataset, metrics, and evaluation protocol to ensure fair comparison. \textbf{Bold} indicates the best.}
\begin{tabular}{lccc}
\toprule
\textbf{Method}
& AUC@5$^\circ$
& AUC@10$^\circ$
& AUC@20$^\circ$ \\
\midrule

SP+SGMNet
& 15.4 & 32.1 & 48.3 \\

SiLK
& 18.0 & 34.4 & 50.4 \\
\midrule
LoFTR
& 22.1 & 40.8 & 57.6 \\

SEEM+SGAM+LoFTR
& 23.4 & 41.8 & 58.7 \\

MESA+LoFTR
& 22.9 & 41.8 & 58.4 \\

\textbf{\method}
 & \textbf{23.89}{\footnotesize +8.1\%} & \textbf{43.48}{\footnotesize +6.6\%} & \textbf{60.74}{\footnotesize +5.5\%} \\

\bottomrule
\end{tabular}
\label{tab:pose_estimation}

\end{table}
\method achieves the best accuracy across all thresholds, reaching AUC@$5^\circ$/$10^\circ$/$20^\circ$ of 23.89/43.48/60.74. It surpasses the LoFTR baseline by 8.1\%/6.6\%/5.5\% and also outperforms the area-level method MESA+LoFTR. Since all methods share the same LoFTR matcher, this gain is directly attributable to the polygon-constrained correspondences provided by \method. The challenging indoor conditions of ScanNet (wide baselines, textureless regions, and repetitive structures) make these consistent improvements across all thresholds particularly meaningful, confirming that polygon-level structural constraints provide geometrically precise inputs for pose estimation.
Notably, \method attains this superior accuracy using far fewer but better spatially distributed correspondences than MESA.
This is attributed to the precise polygon matching produced by LHBGO, which provides mutually non-overlapping spatial constraints for correspondence extraction.
The resulting point correspondences exhibit lower matching redundancy and higher inlier ratios, allowing MAGSAC to converge more efficiently to the correct essential matrix.

\subsubsection{Polygon cloud reconstruction}
As described in Sec.~\ref{sec:polycloud_pipeline}, matched polygons were projected into 3D space to generate polygon clouds.

\begin{figure}[h!t]
    \centering
    \includegraphics[width=0.8\linewidth]{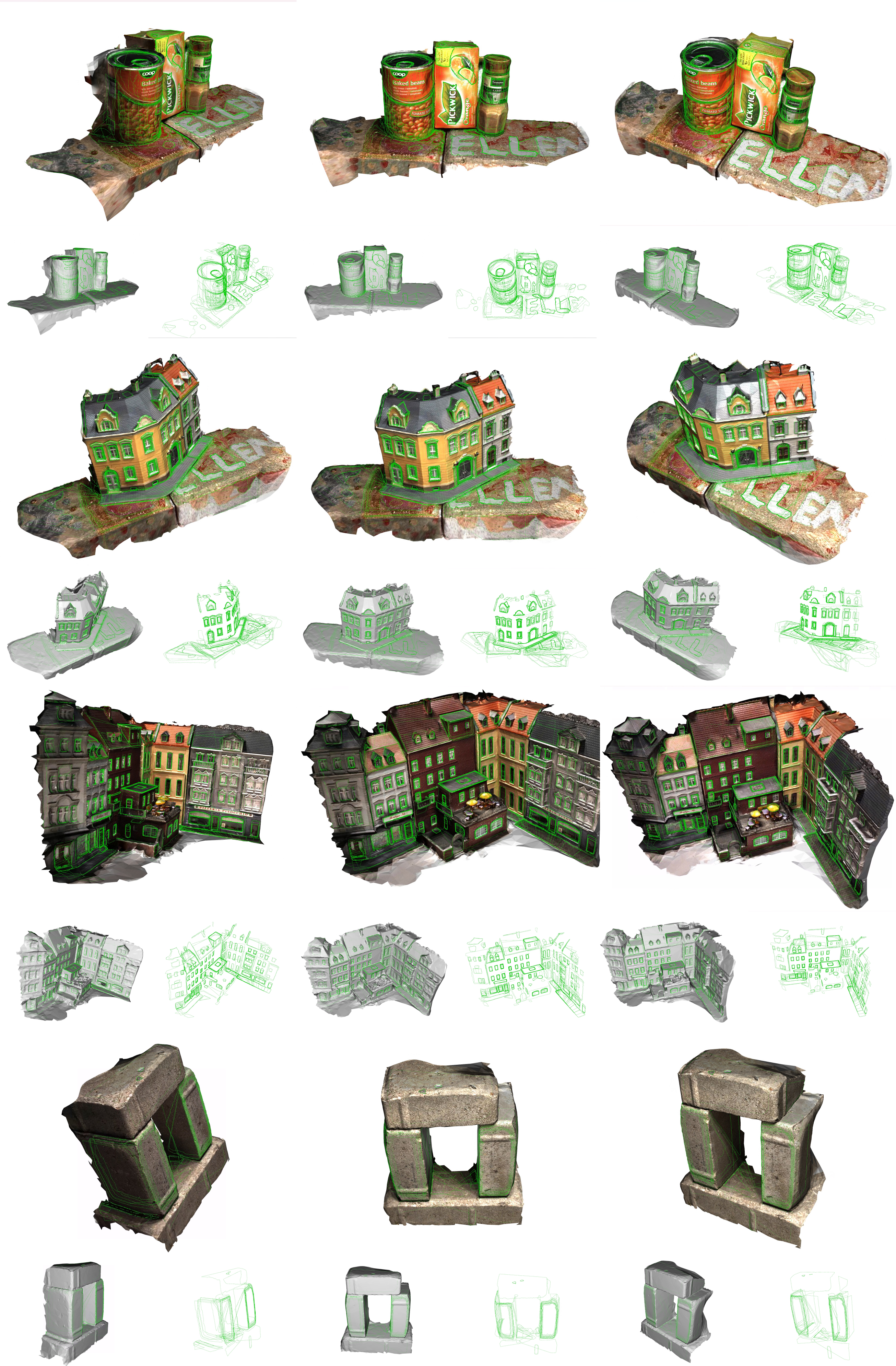}
    \caption{Polygon cloud reconstruction on the DTU dataset. Four scenes are shown, each from three viewpoints. Polygon boundaries (bold contour lines) conform to the 3D surface geometry, providing semantically rich and topologically consistent 3D representations.}
    \label{fig:polygon_cloud}
\end{figure}

Fig.~\ref{fig:polygon_cloud} shows polygon cloud reconstruction results on four scenes from the DTU dataset, each visualized from three viewpoints.
The polygon boundaries remained clearly distinguishable in 3D, demonstrating that the method preserved topologically consistent, surface-faithful boundaries across complex multi-view scenes.

\section{Discussion}
\subsection{Quantity of polygon matches}
Although \method achieves the best performance, the absolute quantity of matched polygons is constrained by the zero-shot pipeline and the complexity of stereo imagery.

\subsubsection{Initial polygon extraction count}
A subset of polygons fails to be extracted during the vectorization due to the performance of the segmentation model, excluding them from subsequent matching.
The capability of the deep-learning-based segmentation model depends on both the sample size and the model parameters \citep{wangOverviewIndustrialImage2025}.
Notably, since \method integrates both texture and geometric matching capabilities, it remains relatively insensitive to upstream segmentation models like SAM.
It ensures robust segmentation performance (Table~\ref{exp_table:sam}), thereby maximizing the quantity of available matches.
As a sequential Detect-then-Match pipeline, \method is inherently subject to error propagation, where such missed or over-segmented polygons may affect downstream matching. Rather than treating the three stages independently, \method adopts a progressive quality-control paradigm in which each stage corrects the residual errors left by the previous one: reliable global matches serve as local anchors, and the local-holistic optimization of LHBGO then re-matches and resolves the hard cases instead of propagating them. The stable precision across SAM backbones (Table~\ref{exp_table:sam}) confirms that this stage-wise correction effectively suppresses error propagation.

\subsubsection{Repeatability of polygon extraction}
Significant viewpoint variations often introduce noise to vector edges, resulting in the same physical object being segmented into distinct shapes.
The LoJoGM module addresses these topologically inconsistent pairs through local-holistic optimization rather than independent thresholding, so that an inconsistent pair is left unassigned only when no candidate within the local subgraph yields a globally consistent assignment.
When the geometric discrepancy between a source polygon and its candidates exceeds the tolerance of the bipartite graph optimization, \method correctly rejects the correspondence rather than forcing a low-quality match.
\subsubsection{Initial matching performance}
Extreme scenarios, such as significant spectral differences or lack of texture, can lead to large errors in initial matching \citep{xuIGEVIterativeMultiRange2025}, ultimately resulting in failure.
If the initial search window drifts significantly due to a lack of distinctive features, the correct target polygon may fall outside the candidate search range.
Although the integration of geometric similarity in \method helps mitigate errors in texture-poor regions, these conditions remain a significant factor limiting the total number of successful matches.

\subsection{Efficiency of BPM on large-format images}
Large-format images refer to high-resolution remote sensing or aerial data (e.g., the ISPRS dataset) \citep{xiaoRTOLLIRobustRealTime2025}, typically exceeding $2000\ \times2000$ pixels and containing macroscopic features.
The challenges in this context are as follows: (1) processing large-format imagery incurs high computational costs, (2) there is a significant relative scale difference between the initial search space and the matching targets (i.e., polygons).

Ablation studies demonstrate that BPM effectively addresses these issues through the following mechanisms: (1) Coarse-to-fine matching.
This strategy establishes a coarse global geometric constraint, drastically reducing the search space.
The initial coarse match provides a necessary disparity prior, determining the search window for each potential polygon correspondence and significantly lowering matching costs.
(2) Adaptive search window.
Matching results from upper levels are projected downward through pyramid layers to progressively refine localization, eliminating the need for exhaustive searches on the full-resolution image.
This approach ensures that the worst-case runtime scales linearly with the number of pyramid levels rather than quadratically with image resolution.

As shown in experiments on the challenging large-format ISPRS dataset (Table~\ref{exp_tabel:5}), bidirectional pyramid matching efficiently handles large-format images.
Unlike other approaches \citep{edstedt2024roma, edstedtDKMDenseKernelized2023}, the BPM matches efficiently polygons of images instead of full-size dense image matching.
\subsection{Robustness of LoJoGM on topological inconsistencies}
While upstream models provide primitives, the resulting greedy correspondence checks are insufficient to resolve topological inconsistencies, where conflict ambiguity arises at object boundaries.
LoJoGM resolves this by reformulating correspondence as a local-holistic bipartite graph optimization, which does not rely on explicit disparity and topology relationships.
Such relationships typically assume topological continuity, but they often tend to break at object boundaries.
Instead, LoJoGM searches for the optimal local overall matching relationship within each subgraph, which allows it to avoid mismatches caused by topological discontinuities.
It is therefore an adaptation mechanism rather than a filtering one: instead of accepting or rejecting each candidate independently against a fixed threshold, LHBGO re-matches the polygons that global matching could not resolve and weighs all local candidates jointly through an augmenting-path search, so that a tentative assignment can be revoked once a better global configuration is found. A correspondence is left unassigned only when no local geometric or textural evidence can accommodate it without degrading the overall solution. This combinatorial assignment perspective is consistent with learning-based matchers such as SuperGlue \citep{sarlinSuperGlueLearningFeature2020} and LightGlue \citep{lindenbergerLightGlueLocalFeature2023}, which likewise cast matching as an optimal assignment problem; the key distinction is that LHBGO confines the optimization to local subgraphs anchored by reliable global matches, which naturally isolates regions of disparity discontinuity without any training.
The ablation results in Table~\ref{exp_tabel:6} confirm this point, showing that the accuracy drops by about 80\% when LoJoGM is replaced by greedy matching.

\subsection{Precision-recall trade-off}
We address this by examining Table~\ref{tab:discrete_exp}: \method maintains a Recall of 72.67\% and 70.92\% on SceneFlow and KITTI respectively, comparable to or exceeding the coverage achieved by competing methods that produce far more raw correspondences.
The Precision values (87.50\% and 82.36\%) confirm that the rejected pairs are genuine mismatches rather than valid correspondences being discarded.

LHBGO does not apply a simple threshold to reject candidates; it solves a combinatorial optimization that jointly considers all source-target pairs within each local subgraph.
A correspondence is rejected ($x_{ij}=0$) only when assigning it would increase the total matching cost of the subgraph, i.e., when no feasible assignment can accommodate it without degrading the overall solution quality.
This differs from filtering, where rejection is determined independently per candidate.
Consequently, the rejected correspondences are those for which no consistent geometric or textural evidence exists within the local context, rather than merely difficult cases.

In practice, high-accuracy sparse matching is preferable to low-accuracy dense matching for downstream tasks such as pose estimation, where even a small proportion of outliers can severely degrade MAGSAC++ convergence and produce erroneous essential matrices.
Concretely, the advantage of \method over MESA originates from a precision-recall co-optimization rather than a precision-only gain: \method retains the superior precision while simultaneously raising recall in regions of disparity discontinuity and abrupt depth change, which the rigid spatial partitioning of MESA inevitably discards. Because LoJoGM can still re-match these regions through its local-holistic combinatorial assignment, the final correspondence set is both clean and broadly distributed, and is the direct cause for \method attaining the highest overall AUC among area-level methods.

\subsection{Downstream task analysis}
\subsubsection{Pose estimation}
LHBGO improves area-level pose estimation by replacing fragile explicit topology and disparity assumptions with local-holistic bipartite graph optimization. In challenging scenes with disparity discontinuities and spatial topology changes, MESA relies on tree-based spatial partitioning to define matching regions, which becomes sensitive when viewpoint changes distort inter-region relationships. By contrast, LHBGO produces mutually non-overlapping polygon correspondences across depth discontinuity boundaries, providing spatially well-distributed homologous points for more accurate pose estimation with fewer but more precise correspondences.
Region-level constraints therefore improve pose recovery only when the matched regions are precise and mutually exclusive \citep{zhangSearchingAreaPoint2024a}; spatial constraint quality matters more than correspondence quantity.

\subsubsection{Polygon cloud}
Line clouds \citep{sugiura3DSurfaceReconstruction2015} extended the 3D representation paradigm from 0D point samples to 1D line segments, demonstrating that incorporating structural connectivity improves scene understanding.
Lines encode local orientation and boundary information that isolated points do not carry, enabling applications such as wireframe reconstruction.
Our polygon cloud further advances this progression from 1D to 2D by introducing closed surface-conforming meshes as 3D primitives.
Polygon cloud integrates and extends the advantages of both point clouds and line clouds in a unified representation: (1) the dense mesh surface of each polygon requires a dense point cloud as its geometric substrate, inheriting the spatial sampling density of point clouds; (2) the polygon boundary is a closed polyline that encloses a 2D region conforming to the true 3D object surface (which can be planar or curved depending on the actual surface geometry), extending the open line segment representation of line clouds to topologically complete boundaries; (3) as a closed 2D surface patch conforming to the actual 3D object surface, polygon cloud carries higher-dimensional geometric and semantic information including area, surface normal fields, and curvature, which 0D and 1D representations inherently do not encode.
Thus, the representation evolves from 0D samples to 1D fragments and then to 2D surface patches with explicit boundaries and instance-level identity.

\subsection{Limitations}
First, as a zero-shot method, \method relies on the performance of pre-trained foundation models (SAM for segmentation, LoFTR for feature matching) without task-specific fine-tuning.
The segmentation granularity and matching robustness of these models jointly determine the upper bound of polygon matching quality.
As foundation models continue to advance in capability and generalization, the performance ceiling of zero-shot approaches is expected to be further raised.

Second, pose estimation on certain challenging ScanNet scenes exhibited relatively high angular errors. These scenes involve wide baselines, textureless surfaces, and repetitive structures, resulting in insufficient correspondences or complete matching failures. Although \method still achieved higher accuracy than other area-level methods under the same conditions, the inherent difficulty of these scenes remains a common challenge across all evaluated approaches.

Third, polygon cloud density and quality are jointly influenced by multiple factors. (1) Although SAM represents the current SoTA in image segmentation, many fine-grained polygons remain undetected, limiting the number of successfully matched polygon pairs and consequently the density of the resulting polygon cloud. As vision foundation models continue to improve in segmentation granularity, denser polygon clouds are expected. (2) Matching failures, which may arise from segmentation quality variations across views, reduce the available polygon pairs for 3D projection. (3) When locating 3D polygon vertices via ray casting, noise and holes in the initial reconstructed mesh can cause projection failures or inaccurate surface localization. Although hole-filling and outlier rejection partially mitigate these effects, completely decoupling polygon cloud quality from reconstruction artifacts remains an open problem for future investigation.

\section{Conclusion}
\label{sec:conclusion}
A zero-shot polygon matching framework \method is proposed by combining automatically learned features from pre-trained models with handcrafted features.
(1) To the best of our knowledge, this work represents the first successful and effective attempt at direct polygon matching for stereo images, extending the feature matching paradigm to a higher semantic level without any training.
(2) We addressed the challenges of scale variation in large-format images through bidirectional pyramid matching, significantly improving efficiency and expandability.
(3) LoJoGM mitigates the effect of local disparity discontinuities, enhancing the overall consistency of the matching results.
(4) \method achieves the best performance with competitive speed on benchmarks in the polygon matching task, with an average matching precision of 87\%.

(5) Two downstream tasks validate the practical utility of polygon matching.
\method achieves the best performance among area-level pose estimation methods despite using fewer correspondences, confirming that low-redundancy polygon-constrained matching provides geometrically precise inputs for photogrammetric applications. Furthermore, we are the first to formally define and construct polygon cloud through the optimal polygon surface generation algorithm, advancing the reconstruction paradigm from point clouds and line clouds to closed surface primitives.

Future work will mitigate the limitations of SAM segmentation and extend polygon matching toward full dense correspondence.

\section*{CRediT authorship contribution statement}
\textbf{Chang Li}: Writing – review and original draft, Conceptualization, Investigation, Methodology, Supervision, Project administration, Funding acquisition, Resources.
\textbf{Xingtao Peng}:  Data curation, Formal analysis,  Software, Validation, Visualization, Writing – editing.

\section*{Acknowledgements}
The authors are grateful for the comments and contributions of the editors, anonymous reviewers and the members of the editorial team. This work was supported by the Key Program of the National Natural Science Foundation of China under Grant Nos. 42030102, the National Natural Science Foundation of China (NSFC) under Grant Nos. 41771493 and 41101407, and the Fundamental Research Funds for the Central Universities under Grant CCNU25JCPT001 and CCNU22QN019.

\section*{Data availability statement}
The dataset that supports the findings of this study is available through the private link: \url{https://figshare.com/s/f0c16d1ff85a99b92c2f}

\bibliographystyle{elsarticle-harv}
\bibliography{main}



\end{document}